\documentclass[]{bytedance}
\usepackage[toc,page,header]{appendix}

\usepackage{minitoc}
\usepackage{amsfonts}
\usepackage{amssymb}
\usepackage{tabularx}
\usepackage{listings}
\usepackage{xcolor}
\usepackage{cancel}

\usepackage{tabulary,multirow,xspace}
\usepackage{fixmath,mathtools,nicefrac,mmstyle}
\usepackage{subcaption}
\usepackage{caption}
\usepackage{wrapfig} % for wrapfigure
\usepackage[misc]{ifsym} % to use \Letter
\usepackage{colortbl}

\usepackage{wrapfig}
\usepackage{multicol}
\usepackage[most]{tcolorbox}
\usepackage{pifont}

\usepackage{graphicx}
\usepackage[export]{adjustbox}
\usepackage{booktabs}
\usepackage{afterpage}

\definecolor{codegreen}{rgb}{0,0.6,0}
\definecolor{codegray}{rgb}{0.5,0.5,0.5}
\definecolor{codepurple}{rgb}{0.58,0,0.82}
\definecolor{backcolour}{rgb}{0.95,0.95,0.92}
\definecolor{boxblue}{RGB}{57,89,163}
\definecolor{boxbluebg}{RGB}{230,237,250} 
\definecolor{myblue}{RGB}{210, 225, 255}

\lstdefinestyle{mystyle}{
    backgroundcolor=\color{backcolour},   
    commentstyle=\color{codegreen},
    keywordstyle=\color{magenta},
    numberstyle=\tiny\color{codegray},
    stringstyle=\color{codepurple},
    basicstyle=\ttfamily\footnotesize,
    breakatwhitespace=false,         
    breaklines=true,                 
    captionpos=b,                    
    keepspaces=true,                 
    numbers=none,                    
    numbersep=5pt,                  
    showspaces=false,                
    showstringspaces=false,
    showtabs=false,                  
    tabsize=2
}
\definecolor{mygray1}{gray}{.95}
\definecolor{mygray2}{gray}{.9}
\definecolor{mygray3}{gray}{.95}
\usepackage{pifont}% http://ctan.org/pkg/pifont

\newlength\savewidth
\newcolumntype{x}[1]{>{\centering\arraybackslash}p{#1pt}}

\newcommand{\app}{\raise.17ex\hbox{$\scriptstyle\sim$}}

\newcommand{\etc}{\textit{etc.}}

\renewcommand{\emph}[1]{\textit{#1}}

\newcommand{\ours}{DreamLite\xspace}

\usepackage{xcolor}
\usepackage{graphicx}
\usepackage{amssymb}
\usepackage{pifont}
\usepackage{floatrow}
\usepackage{amsmath} 
\usepackage{float}
\usepackage{wrapfig}
\usepackage{multirow}
\usepackage{tcolorbox}
\tcbuselibrary{breakable, skins, raster}
\usepackage{listings}
\usepackage{listings}

\definecolor{commentgreen}{rgb}{0.1, 0.4, 0.1}
\definecolor{keywordblue}{rgb}{0.1, 0.1, 0.7}
\definecolor{stringred}{rgb}{0.7, 0.1, 0.1}

\lstdefinestyle{mystyle}{
    commentstyle=\color{commentgreen},
    keywordstyle=\color{keywordblue},   
    stringstyle=\color{stringred},
    basicstyle=\ttfamily\scriptsize, 
    breaklines=true,
    keepspaces=true,
    showstringspaces=false,
    frame=none,                     
    language=Python, 
}

\title{
    \raisebox{-0.2\height}{\includegraphics[height=1.2em]{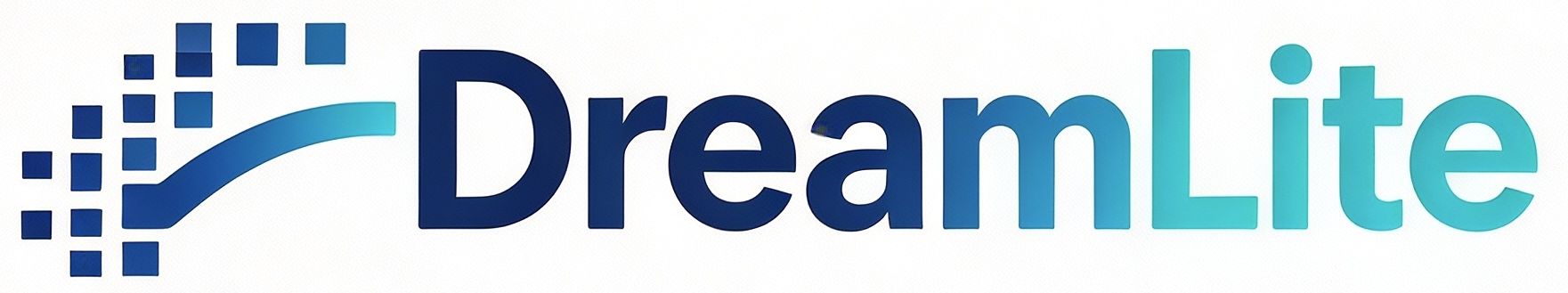}} % 精确控制垂直位移
    : A Lightweight On-Device Unified Model \\
    for Image Generation and Editing \\
}

\author[]{Kailai Feng}
\author[\dagger]{Yuxiang Wei}
\author[]{Bo Chen}
\author[]{Yang Pan}
\author[]{Hu Ye}
\author[]{Songwei Liu}
\author[]{\\Chenqian Yan}
\author[\dagger]{Yuan Gao}

\affiliation[]{Intelligent Creation Lab, ByteDance}
\contribution[\dagger]{Corresponding Author}

\abstract{
Diffusion models have made significant progress in both text-to-image (T2I) generation and text-guided image editing.
However, these models are typically built with billions of parameters, leading to high latency and increased deployment challenges. 
While on-device diffusion models improve efficiency, they largely focus on T2I generation and lack support for image editing.
In this paper, we propose \ours, a \textbf{compact unified on-device diffusion model (0.39B)} that supports both T2I generation and text-guided image editing within a single network. 
\ours is built on a pruned mobile U-Net backbone and unifies conditioning through in-context spatial concatenation in the latent space.
It concatenates images horizontally as input, using a (target $|$ blank) configuration for generation tasks and (target $|$ source) for editing tasks.
To stabilize the training of this compact model, we introduce a task-progressive joint pretraining strategy that sequentially targets T2I, editing, and joint tasks.
After high-quality SFT and reinforcement learning, \ours achieves GenEval (0.72) for image generation and ImgEdit (4.11) for image editing, outperforming existing on-device models and remaining competitive with several server-side models.
By employing step distillation, we further reduce denoising processing to just 4 steps, enabling our \ours could generate or edit a $1024 \times 1024$ image in less than 1s on a Xiaomi 14 smartphone.
To the best of our knowledge, \ours is the first unified on-device diffusion model that supports both image generation and image editing.
}

\date{March 30, 2026}

\checkdata[Project Page]{\url{https://carlofkl.github.io/dreamlite/}}

\begin{document}
\maketitle

\section{Introduction}
\label{sec:intro}

\begin{figure}
    \centering
    \includegraphics[width=\linewidth]{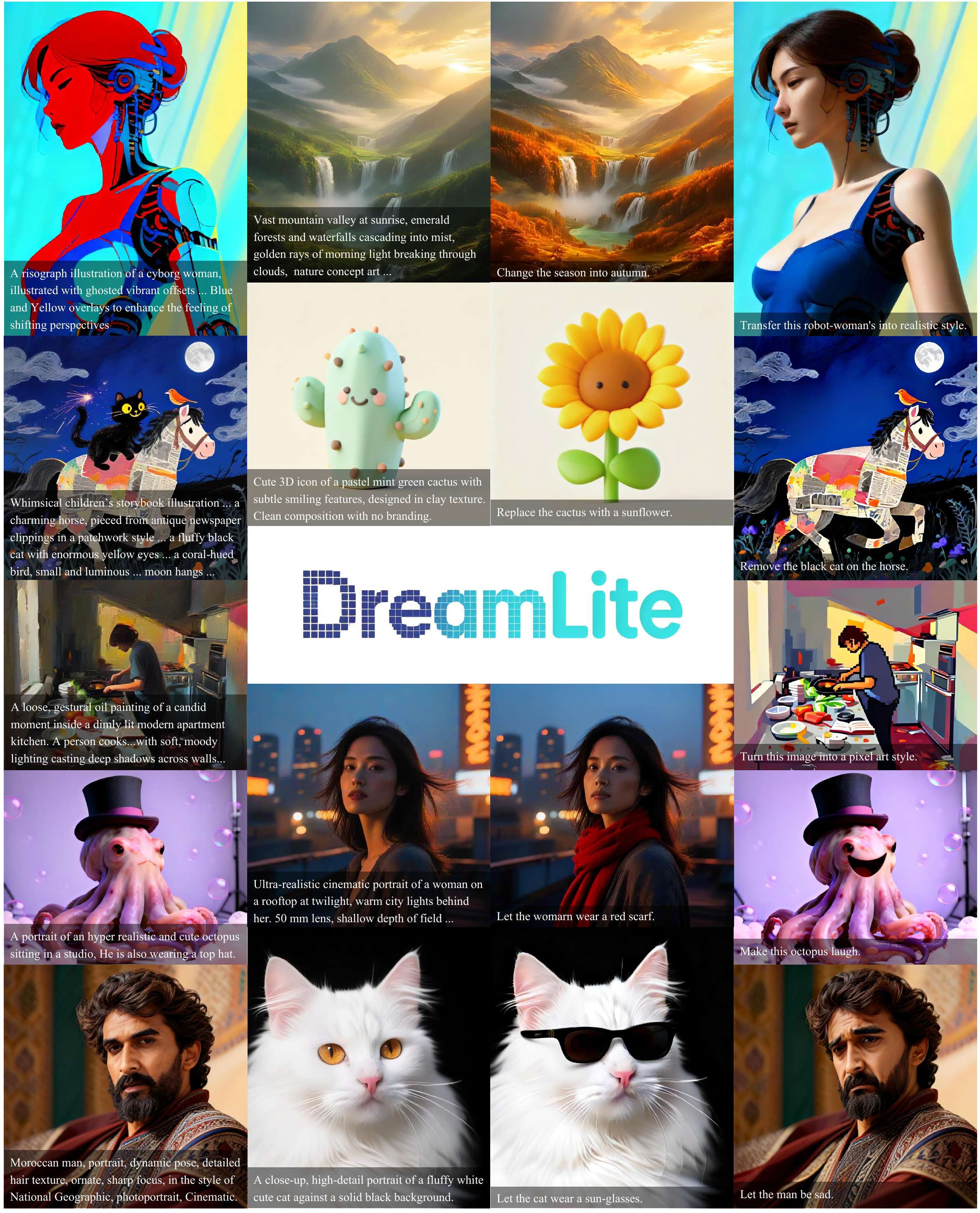}
    % \caption{The proposed \ours enables efficient visual creation on mobile devices, seamlessly supporting both image generation and editing within a unified architecture. As shown, the images on the left are generated by \ours, while the images on the right demonstrate further edits applied to the corresponding left images.}
    \caption{Generation (left) and Editing (right) examples of \ours.}
    \label{fig:open_cover}
\end{figure}

Recent advancements in large-scale diffusion models, such as FLUX series~\cite{flux2023,labs2025flux}, HunyuanImage 3.0~\cite{cao2025hunyuanimage}, Qwen Image~\cite{wu2025qwen} and Seedream series~\cite{seedream2025seedream, gong2025seedream, gao2025seedream} have achieved remarkable progress in both text-to-image (T2I) generation and text-guided editing (I2I). Despite their superior semantic alignment and visual fidelity, these models typically rely on massive backbones with billions of parameters and iterative denoising processes. For instance, FLUX~\cite{labs2025flux} scales its DiT backbone to 12B parameters, imposing prohibitive memory requirements and high inference latency that preclude efficient deployment on consumer-grade devices.
To enhance efficiency, recent research has explored lightweight architectures such as SANA~\cite{xie2024sana}, DeepGen1.0~\cite{wang2026deepgen} and VIBE~\cite{alekseenko2026vibe}, which typically utilize backbones on the order of $\sim$2B parameters. However, achieving stable, real-time performance with these models on mobile hardware remains a significant challenge.

To improve accessibility, several works~\cite{zhao2024mobilediffusion,li2023snapfusion,hu2024snapgen,hu2026snapgen++} focus on deploying compact diffusion models directly on mobile devices. For example, SnapFusion~\cite{li2023snapfusion}, Mobile Diffusion~\cite{zhao2024mobilediffusion} and SnapGen~\cite{hu2024snapgen} leverage lightweight U-Net backbones to achieve quality–efficiency trade-offs for on-device T2I generation. More recently, SnapGen++~\cite{hu2026snapgen++} has explored the potential of Diffusion Transformer for efficient mobile generation. 
However, these approaches predominantly focus on T2I generation and lack support for image editing. In practice, creators demand a unified experience that seamlessly integrates ``generate'' and ``edit'' functionalities within a single application.
Furthermore, deploying two separate models significantly increases system complexity and resource consumption, particularly in memory-constrained devices.

In this paper, we introduce \ours, a unified and compact diffusion model capable of performing both image generation and editing within a single network. 
Following SnapGen~\cite{hu2024snapgen}, we adopt a pruned UNet backbone and extend it for multi-task learning via an in-context conditioning mechanism.
Specifically, we spatially concatenate the target and condition images (left-right) at the input level.
For generation, the target is paired with a blank image; for editing, it is paired with the source image. 
To resolve task ambiguity, we prepend explicit task tokens (\ie, \texttt{[Generate]} or \texttt{[Edit]}) to the text prompts. 
This design enable effective task routing within a shared parameter space without introducing additional parameters or specialized branches.

% Progressive Pretraining?
% 训练
% PT, SFT, RL

Training the compact model with a unified scheme is challenging due to its limited capacity and the divergent optimization objectives of generation and editing tasks.
To ensure stability, we propose a task-progressive joint pretraining scheme.
Specifically, we introduce an intermediate editing pretraining stage between text-to-image pretraining and joint training. 
This stage aligns visual condition with the generative latent space prior to complex unified joint optimization, thereby facilitating stable training.
Consequently, the pretraining of \ours is divided into three progressive stages: \ie, T2I Pretraining $\rightarrow$ Editing Training $\rightarrow$ Unified Joint Training.
Following this, we further adopt a two-stage post-training strategy: supervised fine-tuning (SFT) on a curated high-quality dataset, followed by reinforcement learning (RL) for preference alignment.
During RL training, we employ HPSv3~\cite{ma2025hpsv3} as the reward model for generation and EditReward~\cite{wu2025editreward} for editing, optimizing the diffusion model via ReFL~\cite{xu2023imagereward}. 
This post-training phase consistently enhances both perceptual quality and instruction following, enabling our compact model to outperform prior on-device diffusion baselines.

Overall, \ours achieves GenEval (0.72) and DPG (85.8) for text-to-image generation, along with ImgEdit (4.11) and GEdit (6.88) for text-guided image editing. 
It outperforms specialized lightweight baselines such as SnapGen, SANA-0.6B and VIBE while remaining competitive with larger unified models such as OmniGen2 and Bagel. 
These results validate the efficacy of our unified architecture.  
To minimize deployment overhead, we apply DMD2~\cite{yin2024improved} to compress the sampling process into 4 denoising steps. 
On a representative smartphone (\ie, Xiaomi 14), \ours completes a $1024 \times 1024$ generation or editing task in less than 1 second.
To the best of our knowledge, \ours is the first unified on-device diffusion model to support both generation and editing within a single network for mobile deployment.

Our contributions are summarized as follows:

\begin{itemize}
    \item We propose, to the best of our knowledge, the first unified on-device model that supports both text-to-image generation and text-based image editing, eliminating the need to deploy two separate models.
    \item We introduce an in-context conditioning mechanism for UNet to unify generation and editing, and propose a task-progressive joint pretraining scheme (\ie, T2I → Edit → Unified Joint Training) to stably train the model.
    \item \ours achieves competitive performance on standard benchmarks and consistently outperforms prior mobile models. After deployment on Xiaomi 14, \ours could generate or edit a $1024 \times 1024$ image in less than 1s.
    % \item We will open source our code and model. 
    % \item We introduce an in-context generation paradigm for U-Net diffusion by spatially concatenating target and conditional images, and prepend explicit task tokens for robust task disambiguation. 
    % \item We further adopt a multi-stage training strategy (`T2I → Edit → Unified Joint Training`) to stabilize optimization in small models and improve both generative priors and edit controllability.
    % \item We apply RLHF to boost perceptual quality and instruction following, surpassing existing mobile diffusion baselines. We perform DMD2 distillation to achieve both generation and editing in four steps, enabling efficient deployment and real-world usage on smartphones.
    % \item Our method achieves competitive performance on standard benchmarks (e.g., GenEval: xxx, ImgEdit: xxx) and consistently outperforms prior mobile-oriented diffusion models.
\end{itemize}
\clearpage

\section{Related Work}
\label{sec:rel}

\subsection{Unified Generative Models}

Large-scale image models are increasingly built as unified generative systems that support both text-to-image generation and instruction-based editing through a single model. 
%
%  TODO check一下gpt和gemini的引用
Recent model families such as FLUX 2~\cite{flux2025}, HunyuanImage~\cite{cao2025hunyuanimage}, Seedream 4.0~\cite{seedream2025seedream}, Qwen-Image-2~\cite{wu2025qwen}, as well as commercial systems like Gemini-Image~\cite{GPT-Image}, GPT-Image~\cite{Gemini}, LongCat-Image~\cite{team2025longcat} and DeepGen~\cite{wang2026deepgen}, all move toward ``generate + edit'' as first-class capabilities, typically by strengthening text–image alignment and instruction following with large backbones and curated post-training.
A representative line is FLUX.2, which frames unified generation and editing via an in-context formulation.
Compared to these large or cloud-oriented unified systems, our work targets a \emph{single compact} diffusion model that supports both generation and instruction-based editing for \emph{on-device} deployment under strict memory/latency constraints.

\subsection{Efficient Diffusion Models}
% \vspace{-7pt}
A growing body of work improves diffusion efficiency by optimizing architectures, attention mechanisms, and training/inference recipes.
PixArt-$\Sigma$~\cite{chen2024pixart} explores transformer efficiency for high-resolution generation with key-value compression to alleviate attention cost at large token counts.
And SANA~\cite{xie2024sana} proposes linear-attention diffusion transformers to reduce the quadratic complexity of self-attention at high resolutions while maintaining competitive generation performance.
Beyond pure generation, EditMGT~\cite{chow2025editmgt} and VIBE~\cite{alekseenko2026vibe} presents a compact instruction-based editing pipeline that combines a lightweight vision-language model with an efficient diffusion backbone, demonstrating that strong editing can be achieved without relying on extremely large diffusion models.
Our method is complementary to these efforts: rather than only accelerating T2I or only optimizing editing, we focus on an efficient \emph{unified} interface that consolidates T2I and editing into a single compact model.
%
% Recent efforts aim to improve diffusion efficiency. PixArt-$\Sigma$~\cite{chen2024pixart} introduces key–value compression for 4K image generation, while SANA employs linear self-attention to enable efficient synthesis on consumer GPUs. 
% %
% VIBE xxx

% 一些加速的方法要不要算进来，
% \vspace{-7pt}
\subsection{On-Device Generative Models}
% \vspace{-7pt}
To enable on-device deployment, prior works have explored quantization, pruning, and knowledge distillation to reduce model size and latency. 
Early on-device systems~\cite{li2023snapfusion, zhao2024mobilediffusion} pruned and distilled U-Net architectures to generate 512-pixel images within seconds. 
SnapGen~\cite{hu2024snapgen} demonstrated that a compact U-Net derived from SDXL can generate $1024 \times 1024$ images on mobile devices with a carefully engineered architecture and training recipe. 
More recently, SnapGen++~\cite{hu2026snapgen++} explores efficient diffusion transformers tailored for mobile and edge deployment.
Concurrent to our work, Mobile-O~\cite{shaker2026mobile} attempts to unify both visual generation and understanding within a single compact framework. 
However, since Mobile-O relies on an understanding-centric paradigm to execute generation tasks, it struggles with fine-grained visual control and spatial consistency at in editing tasks. 
Consequently, its performance in complex image editing scenarios remains somewhat suboptimal. 
Despite this progress, most on-device works primarily emphasize T2I generation, while instruction-based editing often requires either separate models or additional editing-specific components. 
Our work targets this gap by providing a unified interface and training strategy so that a single compact model supports both ``generate an image'' and ``edit my photo,'' reducing deployment complexity and resource consumption on mobile devices.

% \vspace{-7pt}
\subsection{RLHF}
% \vspace{-7pt}
Post-training alignment has emerged as a crucial stage for enhancing perceptual quality and instruction compliance, surpassing the limitations of noisy web-scale pre-training.
A common practice involves utilizing learned reward models (\eg, ImageReward~\cite{xu2023imagereward}, HPSv2/v3~\cite{wu2023human, ma2025hpsv3}, PickScore~\cite{kirstain2023pick}, and the editing-specific EditReward~\cite{wu2025editreward}) as optimization targets to improve aesthetics and prompt faithfulness while mitigating visual artifacts.
On the optimization front, various methods adapt reinforcement learning (RL) to generative models. 
For instance, ReFL~\cite{xu2023imagereward} performs reward-guided fine-tuning for diffusion models under constrained backprop settings to reduce training cost, and has inspired follow-up explorations of more efficient preference optimization.
Parallel to RL-based approaches, DPO-like objectives (\eg, Diffusion-DPO~\cite{wallace2024diffusion}, AlignProp~\cite{prabhudesai2023aligning}) have been developed to leverage pairwise preferences without explicit reward modeling.
Recent flow-model works also explore GRPO variants (\eg, Flow-GRPO~\cite{liu2025flow}, DanceGRPO~\cite{xue2025dancegrpo}) for better stability and instruction following under preference feedback.
% 
% Additionally, the use of vision-language models as "AI Feedback" (RLAIF), exemplified by VLM-Reward~\cite{baumli2023vision}, has further automated the alignment process for open-ended generation tasks.

% REFL~\cite{xu2023imagereward},Flow-GRPO~\cite{liu2025flow}, DanceGRPO~\cite{xue2025dancegrpo}, Duffusion-DPO~\cite{wallace2024diffusion}, DiffusionNFT~\cite{zheng2025diffusionnft}

\subsection{Step Distillation}

Few-step sampling is essential for interactive and on-device applications. 
A representative line is consistency-based distillation, including Latent Consistency Models (LCM)~\cite{luo2023latent}, which distill time-consistent behavior to enable generation in a small number of steps. 
Another influential family is distribution matching distillation.
DMD~\cite{yin2024one} distills a multi-step teacher into a few-step student via distribution matching objectives, and DMD2~\cite{yin2024improved} further improves stability and sample quality under aggressive step reduction. 
In practice, these methods are often used to compress sampling to $\le 4-8$ steps with acceptable quality, and have been integrated into various backbones (including SD, SDXL and QwenImage, \etc). 
Adversarial-based distillation is also a prominent approach; for instance, SDXL-Turbo and SD-Turbo~\cite{sauer2024adversarial} utilize Adversarial Diffusion Distillation (ADD) to achieve high-fidelity, real-time synthesis in a single step.
Recent efforts such as RG-LCD~\cite{li2024reward} and DI++~\cite{luo2024diff} incorporate reward model into the distillation process with an additional score model to maintain proximity to the original generator. To address this, LaSRO~\cite{jia2025reward} optimizes arbitrary rewards via latent space exploration. Reward-Instruct~\cite{luo2025reward} aligns generators with rewards without requiring training images. TAFS-GRPO~\cite{yue2026know} eliminates the need for differentiable reward functions by leveraging a policy gradient algorithm.
In this work, we employ the DMD2 to compress our sampling process to 4 steps.
% 
% Flow-based distillation has also gained attention alongside the rise of Flow Matching. Methods like SwiftBrush~\cite{nguyen2024swiftbrush} and Rectified Flow~\cite{liu2022flow} (\eg, in SD3) simplify sampling trajectories into straight lines, facilitating efficient few-step inference through Euler integration or post-training refinement.

% LCM~\cite{luo2023latent}, DMD~\cite{yin2024one}, DMD2~\cite{yin2024improved}, xxx

\section{Method}
\label{sec:method}

\begin{figure*}[t]
    \centering
    \includegraphics[width=\linewidth]{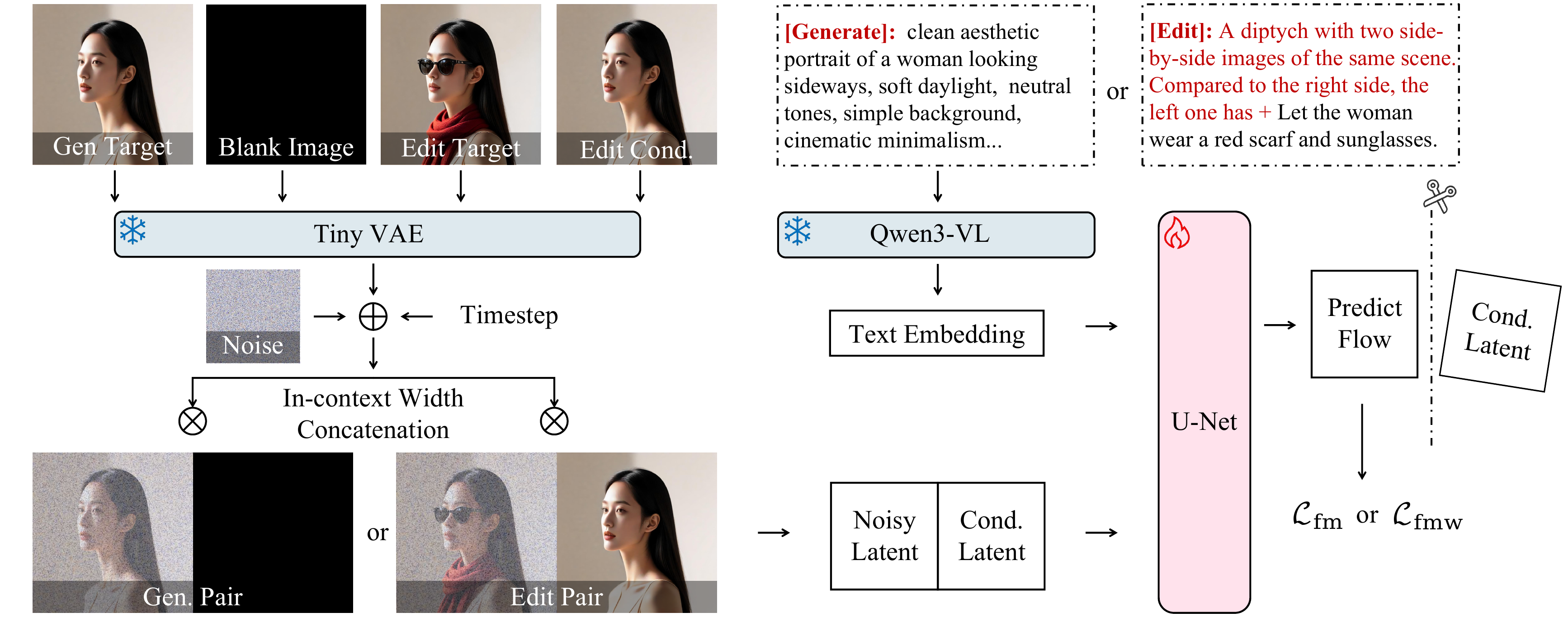}
    \caption{Overview of \ours architecture. It consists of three primary modules: a UNet backbone, a Variational Autoencoder (VAE), and a text encoder. To support unified generation and editing, an in-context conditioning framework is adopted that unifies both tasks at the input level.}
    \label{fig:arch}
\end{figure*}

% \begin{table*}[t]
%     \centering
%     \caption{Architectural evolution of \ours. We calculate the Parameters (Params) and GFLOPs of the U-Net.}
%     \label{tab:arch_evolution}
%     \resizebox{0.9\textwidth}{!}{
%         % \setlength{\tabcolsep}{6pt} 
%         \begin{tabular}{llcc}
%             \toprule
%             \textbf{Stage} & \textbf{Optimization Strategy} & \textbf{Params (M)} & \textbf{FLOPs (G)} \\
%             \midrule
%             A & SDXL Baseline + Text Token Refiner & 2567.46 & 3054.08 \\
%             B & Pruning (Blocks $\rightarrow [0, 2, 4]$; Dimensions $\rightarrow [256, 512, 896]$) & 839.73 & 1084.51 \\
%             C & Remove Self-Attention in High-Resolution Stages & 829.24 & 1041.56 \\
%             D & Replace Conv. with Expanded Separable Conv. & 719.38 & 748.86 \\
%             E & Trim FFN Layers (expansion ratio: $4 \rightarrow 3$) & 653.66 & 657.46 \\
%             F & Replace MHSA with Multi-Query Attention (MQA) & 517.74 & 600.79 \\
%             G & Stage Alignment (Dwon\&Up Stage: $3\rightarrow2$, Mid Block: 6) & {382.99} & {579.90} \\
%             H & Add QK-RMSNorm, Light Text Projector & {389.97} & {607.60} \\
%             \bottomrule
%         \end{tabular}
%     }
% \end{table*}

\begin{figure*}
    \centering
    \includegraphics[width=\linewidth]{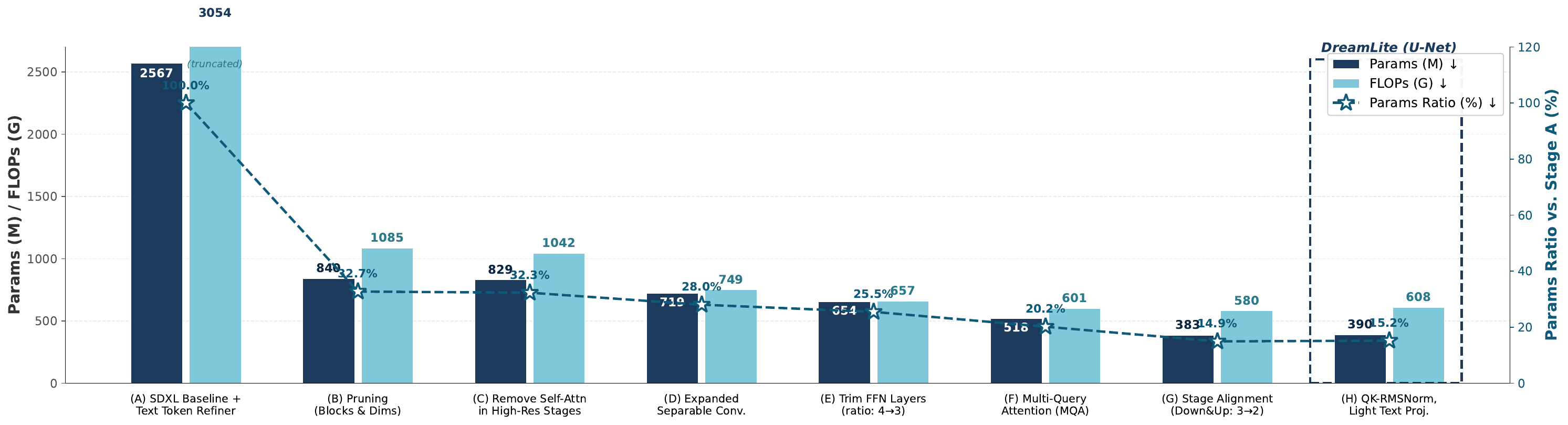}
    \caption{Architectural evolution of \ours. We calculate the Parameters (Params) and GFLOPs of the U-Net.}
    \label{fig:arch_evolution}
\end{figure*}

This section presents the training pipeline and details of our \ours.
We first describe the model architecture (Section~\ref{sec:method_arch}), and then detail our training procedure, including task-progressive joint pretraining (Section~\ref{sec:method_pretrain}), post training (Section~\ref{sec:method_post}), and few-step distillation (Section~\ref{sec:method_distill}).

\subsection{Model Architecture}
\label{sec:method_arch}

As illustrated in Fig.~\ref{fig:arch}, the architecture of \ours consists of three primary modules: a UNet backbone, a Variational Autoencoder (VAE), and a text encoder.
To support unified image generation and editing, we further introduce the in-context conditioning mechanism.

\noindent \textbf{Variational Autoencoder.}
Following~\cite{rombach2022high, flux2023}, we adopt a latent diffusion framework. 
To enable efficient on-device deployment, we employ an extremely lightweight VAE (\ie, TinyVAE~\cite{taesd2024}), which contains only 2.5M parameters for image tokenization. 
It maps image $x$ as a 4-channel latent $z$ with an $8\times8$ downsampling factor, facilitating efficient training and inference.

\noindent\textbf{Compact UNet.}
Efficiency is the primary objective guiding the design of \ours. 
To this end, we build upon the mobile-efficient T2I architecture of SnapGen~\cite{hu2024snapgen}, a systematically compressed version of SDXL~\cite{podell2023sdxl}. 
Specifically, we optimized the U-Net backbone by making it both shallower and thinner: the number of transformer blocks was reduced from [0, 2, 10] to [0, 2, 4], and the channel dimensions were shrunk from [320, 640, 1280] to [256, 512, 896] and a latent sample size is $128 \times 128$.
We further enhanced efficiency through several key optimizations: 
\begin{itemize}
    \item  Remove self-attention layers at high-resolution stages to mitigate quadratic complexity;
    \item Replace standard convolutions with expanded separable convolutions (\ie, depthwise convolution \& pointwise convolution);
    \item Set the hidden channel expansion ratio to $3$ in the feed-forward network.
    \item Adopt Multi-Query Attention (MQA) with a single KV head to reduce both computational overhead and memory footprint;
    \item Stage alignment, add QK-RMSNorm and a light text projector.
\end{itemize}

As summarized in Fig.~\ref{fig:arch_evolution}, this step-by-step optimization successfully compresses the 2.5B baseline into a highly efficient 389M parameter backbone, significantly reducing FLOPs while preserving generative performance.
Additional architectural details can be found in the original SnapGen paper~\cite{hu2024snapgen}.

\noindent \textbf{Text Encoder.}
For text conditioning, we utilize Qwen3-VL-2B~\cite{bai2025qwen3} as our text encoder. 
We leverage its robust visual-language comprehension capabilities to accurately interpret complex user instructions and process multimodal inputs. 
This choice ensures precise semantic alignment between the input instructions and the generated content.

\noindent\textbf{In-context Paradigm.}
Existing UNet-based image editing methods~\cite{geng2024instructdiffusion,brooks2023instructpix2pix,huang2024smartedit} typically follow the InstructPix2Pix paradigm~\cite{brooks2023instructpix2pix}, which concatenates the condition image with the noisy latent in the channel dimension and fine-tunes the model. 
However, this mechanism inevitably degrades the generative priors of the pretrained text-to-image (T2I) model and hinders the development of a unified architecture.
To address these limitations, we propose extending the UNet with an in-context conditioning framework that unifies both image generation and editing tasks at the input level within a single compact network.
As shown in Fig.~\ref{fig:arch}, we construct a two-panel latent by concatenating the latent of the target image ($z_{\text{tgt}}$) and the conditioning image $z_{\text{cond}}$ along the width (spatial) dimension: 
\begin{equation}
z^{\text{pair}} = \mathrm{Concat} \big(z_{\text{tgt}},\, z_{\text{cond}}\big),
\end{equation}
where the left panel corresponds to the target output and the right panel provides the visual condition.
The concatenated latent $z^{\text{pair}}$ is fed directly into the U-Net.
For text-to-image generation, we set the conditioning panel to a blank (all-black) image $x_{\text{blank}}$ (with latent $z_{\text{blank}}$) to represent ``no visual condition''. 
For image editing, we use the source image $x_{\text{src}}$ (latent $z_{\text{src}}$) as the condition. 
This design allows the model to extend from T2I to editing directly without introducing additional modules, making it highly suitable for a unified framework.
To further reduce task ambiguity when training a single model for two behaviors, we prepend explicit task tokens to the text prompt: \texttt{[Generate]} for generation task and \texttt{[Edit]} for editing task. 
These tokens act as lightweight routing signals without requiring extra parameters or task-specific branches, thereby improving both generation quality and edit controllability under a unified framework.
We also compare this in-context formulation with InstructPix2Pix; further details and motivation regarding the architectural design are provided in Section~\ref{sec:exp}.

\subsection{Task-progressive Joint Pretraining}
\label{sec:method_pretrain}

Training a compact model with a unified formulation is challenging due to its limited capacity and the divergent optimization objectives of generation and editing tasks.
To ensure stable convergence, we propose a task-progressive joint pretraining scheme.
Unlike standard approaches that transition directly from text-to-image pretraining to joint training, we introduce an intermediate editing pretraining stage. 
This stage serves to align the visual conditioning representations with the generative latent space before the complex joint optimization, thereby mitigating task interference.
Specifically, the pretraining of \ours is divided into three progressive stages: (i) T2I Pretraining, (ii) Editing Training, and (iii) Unified Joint Training.

\subsubsection{Text-to-Image Pretraining}
We first train the \ours as a standard text-to-image diffusion model using the flow matching objective~\cite{lipman2022flow,liu2022flow}.
During training, the noisy latent $z_t$ is constructed through linear interpolation between the initial Gaussian noise $\epsilon$ and the clean image latent $z$, \ie, $z_t = F(z, t) = t \cdot z + (1 - t) \cdot \epsilon$ and $t \in [0, 1]$. 
The model is then trained to predict the velocity of the vector field that defines the trajectory between the noise and data distributions, \ie, $v_t = z - \epsilon$. 
The training objective can be formulated as:
\begin{equation}
\mathcal{L}_{\text{fm}} = \mathbb{E}_{t, z, \epsilon, y} \left[ \Vert v_\theta(z_t, t, y) - (z - \epsilon) \Vert^2 \right], 
\label{eq:fm}
\end{equation}
where $v_\theta$ denotes the denoising UNet, $\theta$ represents the learnable parameters and $y$ denotes the conditional embeddings. 
To improve convergence and stability,  we adopt a progressive resolution curriculum following prior work~\cite{hu2024snapgen}. 
The training proceeds sequentially from 256 $\times$ 256 to 512 $\times$ 512, and finally to 1024 $\times$ 1024 resolution, with a multi-scale training strategy applied at each stage. 
Furthermore, following Stable Diffusion 3~\cite{esser2024scaling}, we employ a logit-normal noise sampler to concentrate training on intermediate timesteps.
We also utilize dynamic time shifting~\cite{flux2023} to scale noise levels according to image resolution.
Collectively,  this stage establishes a strong generative prior for subsequent training stages.

\subsubsection{Edit Pretraining}

Following T2I pretraining, we activate the in-context conditioning mechanism and continue training the model on paired text-guided image editing data. 
The primary objective of this stage is to align the newly introduced visual conditioning with the pre-trained generative latent space, enabling the model to generate images based on both condition images and editing instructions.
We also employ a flow-matching loss during this phase.
However, a major challenge in training for editing tasks is that the target edit regions in the target images are usually small.
If a standard uniform loss is applied to the entire image, the gradient signal from these small edited regions can be overwhelmed by the dominant signal from the unchanged background, leading to unsatisfactory editing results.
To mitigate this imbalance, following UniWorld-V1~\cite{lin2025uniworld}, we apply a foreground-emphasis mask to re-weight the loss.
Concretely, for localized editing tasks, we derive the edited region mask through a four-step pipeline:
\begin{itemize}
    \item {Pixel-wise Differencing:} we compute the absolute difference between source and target images, applying a tolerance threshold to identify candidate regions;
    \item {Dilation:} a dilation operator is applied to reduce pixel-level noise; 
    \item  {Connected Component Filtering:} we filter out small, isolated components to eliminate spurious artifacts;
    \item {Max-pooling Downsampling:} max-pooling is utilized to remove internal noise within connected regions and we estimate the edited area size $A_{\text{edit}}$.
\end{itemize}
To prevent small edited regions from being overwhelmed by the static background, we assign higher loss weights to the masked pixels based on the area ratio $x = A_{\text{total}} / A_{\text{edit}}$, where $A_{\text{total}}$ denotes the full image area.
Subsequently, we use a logarithmic weighting function $w(x) = \log_2(x) + 1$ to balance training stability with sensitivity to minor edits. 
This mask strategy is applied exclusively to local editing; for global editing or style transfer, we maintain uniform weighting to preserve global distribution alignment. 

Finally, we compute this weighting mask $w$ to applied to the flow matching objective:
\begin{equation}
\mathcal{L}_{\text{fmw}} = \mathbb{E}_{t, z, \epsilon, y} \left[ \Vert w \odot (v_\theta(z_t, t, y) - w \odot (z - \epsilon)) \Vert^2 \right].
\label{eq:masked_fm}
\end{equation}
Intuitively, the mask focuses the learning signal on regions that differ between source and target images, encouraging the model to make stronger, more faithful edits while preserving the unchanged background.

\subsubsection{Unified Pretraining}
Subsequently, we perform unified joint training on a mixture of T2I and editing data.
This stage is designed to consolidate the generative priors established in the T2I stage with the instruction-following capabilities acquired during Edit pretraining, ensuring the model converges to a single set of parameters that reliably supports both behaviors.
As stated in Sec.~\ref{sec:method_arch}, to mitigate task conflict in this unified setting, we prepend explicit task tokens to the text prompts: \texttt{[Generate]} to indicate the absence of a visual reference for generation tasks, and \texttt{[Edit]} for editing instructions.
These tokens function as lightweight routing signals, guiding the model to switch behaviors dynamically based on the input context. 
Consequently, \ours achieves robust performance in both generation and editing under a strictly unified architecture, eliminating the need for additional parameters.
%
% bucket training,
% To support multi-scale image generation and editing with capabilities up to 2K resolution, we adopt a bucket training strategy\cite{podell2023sdxl,chen2023pixartalpha}.
% % 
% Specifically, We identify 20 representative aspect ratios based on the statistical distribution of our training data. 
% % 
% During training, images are grouped into corresponding buckets to maintain their native geometric proportions and prevent compositional distortion.
% % 
% As illustrated in Fig.~\ref{fig:open_cover}, \ours exhibits robust synthesis quality across a wide spectrum of aspect ratios.

\subsection{Post Training}
\label{sec:method_post}

While our task-progressive joint pretraining establishes a strong foundation for both generation and editing, the model's behavior remains unstable due to the high variance of the pretraining data. 
To improve stability and performance, we employ a two-stage post-training strategy: Supervised Fine-Tuning (SFT) and Reinforcement Learning (RL).

\subsubsection{Supervised Fine-Tuning (SFT).}

The primary goal of SFT is to refine the model's behavior by exposing it to a curated distribution of high-quality data. 
To achieve this, we construct a dataset comprising approximately 0.5M samples, selected for their high visual quality and caption diversity.
By fine-tuning on these cleaner, denser signals, we effectively steer the model's target distribution toward a manifold of higher realism and precise instruction following.

\subsubsection{Reinforcement Learning (RL).}
\label{sec:method_rl}

To further align \ours with human preferences, we incorporate Reinforcement Learning from Human Feedback (RLHF). 
Specifically, we adopt the Reward Feedback Learning (ReFL) framework~\cite{xu2023imagereward}, which directly leverages scalar signals from a pre-trained reward model to guide the denoising trajectory via gradient backpropagation.
During training, given a condition $c$ and a timestep $t$, the model progressively denoises the latent to time $t$ to predict the corresponding clean image $\hat{x}$.
A scalar reward $r(c, \hat{x})$ is then computed using the reward model, and gradients are backpropagated through the denoising process.
We adopt a ReLU-truncated reward formulation:
\begin{equation}
\mathcal{L}_{\text{rl}} = - \max(0, r(c, \hat{x}) - b),
\end{equation}
where $b$ is a hyperparameter introduced to ensure training stability and mitigate reward hacking.
In our experiments, we employ task-specific reward models to optimize distinct capabilities. 
For text-to-image generation, we utilize HPSv3~\cite{ma2025hpsv3}, a state-of-the-art preference model built upon the Qwen-VL backbone, with $b = 11$.
For image editing, we employ EditReward~\cite{wu2025editreward}, which is explicitly designed to evaluate adherence to editing instructions, and use $b=2.5$. 
This post-training stage consistently enhances both perceptual quality and instruction following, enabling our compact model to outperform prior mobile diffusion baselines.

\vspace{-10pt}
\subsection{Step Distillation}
\label{sec:method_distill}

While our post-training strategy ensures high-quality generation, it still requires tens of denoising steps to generate/edit an image, creating a significant bottleneck for real-time mobile applications.
To bridge the gap between high-fidelity synthesis and low-latency deployment, we apply Distribution Matching Distillation (DMD)~\cite{yin2024one,yin2024improved} to reduce the sampling process to just 4 steps without compromising visual quality.
DMD distills a multi-step diffusion teacher into a few-step generator $G$ by minimizing the approximate Kullback-Leibler (KL) divergence between the distribution of real images $p_{\text{real},t}$ and the generator's output distribution $p_{\text{fake},t}$.
The gradient of DMD loss is computed as:
\begin{small}
\begin{equation}
    \begin{aligned}
    \nabla\mathcal{L}_\text{DMD} &=
        \mathbb{E}_t\left(\nabla_\theta \text{KL}(p_{\text{fake},t} \| p_{\text{real},t}) \right) \\ &=
        - \mathbb{E}_t\left(\int \big(s_{\text{real}}(F(G_{\theta}(\epsilon), t), t) - s_{\text{fake}}(F(G_{\theta}(\epsilon), t), t)\big) \frac{dG_\theta(\epsilon)}{d\theta} \hspace{.5mm} d\epsilon\right),
    \end{aligned}
    \label{eq:kl-grad}
\end{equation}
\end{small}
where $\epsilon \sim \mathcal{N}(0,\mathbf{I})$ is a random Gaussian noise, and $\theta$ denotes the generator parameters.
$F$  defines the forward diffusion process that transfers the generated sample $G_\theta(\epsilon)$ the noise level corresponding to timestep $t$.
$s_{\text{real}}$ and $s_{\text{fake}}$ represent the scores, approximated using diffusion models $v_{\text{real}}$ and $v_{\text{fake}}$, respectively.
To stabilize training, we also employ the GAN loss $\mathcal{L}_{\text{GAN}}$~\cite{yin2024improved} to enhance the diversity and realism of the generated images for the distilled model.
Our distilled model enables high-quality image generation and editing in only 4 sampling steps without the need for CFG, showcasing exceptional efficiency.
%
% This efficiency translates to a runtime of ∼\sim∼1s on mobile hardware, significantly outperforming existing solutions.

%
% Specifically, a classification branch is added on top of the bottleneck of the fake diffusion denoiser.
% %
% The classification branch and upstream encoder features in the UNet are trained by maximizing the standard non-saturing GAN objective:
% %
% \begin{equation}
% \label{eq:gan_loss}
% \mathcal{L}_{\text{GAN}} = \mathbb{E}_{x \sim p_{\text{real}}, t \sim [0, T]}[\log D(F(x, t))] + \mathbb{E}_{z \sim p_{\text{noise}}, t \sim [0, T]}[-\log(D(F(G_\theta(z), t)))],
% \end{equation}
% %
% where $D$ is the discriminator, and $F$ is the forward diffusion process (\ie, noise injection), with noise level corresponding to time step $t$.
% %
% The generator $G$ minimizes this objective.

\section{Experiments}
\label{sec:exp}
\subsection{Implementation Details}
\ours is implemented using the PyTorch. 
The training corpus consists of 20M text-to-image pairs and 1.7M image editing samples.
During the unified joint training stage, we employ a sampling ratio of approximately 1:1 between T2I and editing data to balance performance. 
We utilize the AdamW optimizer with a staged learning rate schedule: $1 \times 10^{-4}$ for initial T2I pre-training, $1\times10^{-5}$ for editing training and $1\times10^{-6}$ for unified joint training. 
The batch size is set to 576.
%
% More details are provided in the Supplementary Material.
% 
% For alignment, we utilize the (xxx RLHF, penalty of xxx). 
% 
% Finally during the DMD phase, we use the pre-trained \ours as the teacher model to facilitate 4-step inference.
\subsection{Dataset Information}

\begin{table}[t]
    \centering
    \small % 略微缩小字体以适配内容
    \caption{Detailed distribution of the 21.7M total samples.}
    \label{tab:dataset}
    \resizebox{0.7\textwidth}{!}{
        \begin{tabular}{llr}
        \toprule
        \textbf{Category} & \textbf{Source / Content Type} & \textbf{Samples} \\
        \midrule
        \multicolumn{3}{l}{\textit{Part A: Generation Data (20.0M)}} \\
        \midrule
        General Perception & COYO, LAION, JoourneyDB, \etc. & 14.8M \\
        Human \& Portrait & High-quality human, Portrait subsets & 0.4M \\
        Graphic Design & Text content, Canva subsets & 0.4M \\ 
        Scene Text & Dense/Large scene text & 1.5M \\
        {Artistic Styles} & Midjourney style prompts & 0.6M \\
        {Specialized Prompts} & Object relations, Text rendering & 2.4M \\
        \midrule
        \multicolumn{3}{l}{\textit{Part B: Editing Data (1.74M)}} \\
        \midrule
        Understanding & Feature Extraction \& Parsing, Action & 429k \\
        Local Edit & Add, Remove, Color, Replace, \etc. & 830k \\
        {Global Edit} & Enhance, Lighting, Background & 300k \\
        {View Edit} & Camera View, Zoom In/Out, Outpainting & 151k \\
        {Style Edit} & Unreal $\leftrightarrow$ Real & 31k \\
        \bottomrule
        \end{tabular}
    }
\end{table}

\noindent\textbf{Text-to-image Generation Dataset.} 
To establish a robust and diverse generative prior, our text-to-image (T2I) pre-training utilizes a curated corpus of approximately 20M samples. 
We categorize these data into five primary domains: General Perception, Human \& Portrait, Graphic Design, Scene Text, and Artistic Styles. 
Specifically, we leverage large-scale open-source datasets (\eg, LAION~\cite{schuhmann2022laion}, COYO~\cite{kakaobrain2022coyo}, JourneyDB~\cite{sun2023journeydb}) for general semantic alignment, while incorporating specialized subsets for high-quality human generation and complex text rendering. 
For all samples, we prioritize high-resolution images and apply a stringent filtering pipeline to remove low-quality captions samples. 
The data composition is summarized in Table~\ref{tab:dataset}.

\noindent\textbf{Image Editing Dataset.} 
To empower \ours with comprehensive instruction-following capabilities, we curated an image editing corpus comprising approximately {1.7M samples}. 
These samples are categorized into five primary groups:
\begin{itemize}
    \item {Understanding Edit:} A significant portion of the data is for editing-oriented understanding such as feature extraction and human action adjustments (\eg, age, expression, pose).
    \item {Local Edit:} This includes fine-grained modifications such as object addition/removal and color/material changes, \etc.
    \item {Global Edit:} This category focuses on overall image enhancements, including lighting direction adjustments, background modifications, \etc.
    \item {View Edit:} It involves spatial and structural transformations, such as camera view adjustments, zooming (in/out) and object-level rotation or motion.
    \item {Style Edit:} This one mainly focus on style transfer (unreal-to-real).
\end{itemize}
The detailed data distribution across these fine-grained categories is presented in Table~\ref{tab:dataset}.

\subsection{Quantitative Results}
To evaluate the performance of \ours, we conduct extensive comparative experiments across multiple mainstream benchmarks, focusing on both Image Editing and Image Generation capabilities.

\begin{table*}[t]
\centering
\caption{
Evaluation on GenEval benchmark. All images are generated at $1024 \times 1024$ resolution. ``Param.'' denotes backbone parameters.
}
\label{tab:geneval_results}
\resizebox{\textwidth}{!}{
    \begin{tabular}{lc|cccccc|c}
    \toprule
    % \multirow{2}{*}{\textbf{Model}} & \textbf{Backbone} & \multicolumn{7}{c}{\textbf{GenEval}} \\ & \textbf{Params} & Single Obj. & Two Obj. & Counting & Colors & Position & Color Attr. & Overall \\ 
    {Model} & {Param.} & Single Obj. & Two Obj. & Counting & Colors & Position & Color Attr. & Overall \\
    \midrule
    FLUX.1-Dev~\cite{flux2023} & 12B & 0.99 & 0.81 & 0.79 & 0.74 & 0.20 & 0.47 & 0.67 \\
    BAGEL~\cite{deng2025bagel} & 7B & 0.99 & 0.94 & 0.81 & 0.88 & 0.64 & 0.63 & 0.82 \\
    OmniGen2~\cite{wu2025omnigen2} & 4B & 1.00 & 0.95 & 0.64 & 0.88 & 0.55 & 0.76 & 0.80 \\
    Z-Image~\cite{cai2025z} & 6B & 1.00 & 0.94 & 0.78 & 0.93 & 0.62 & 0.77 & 0.84 \\
    LongCat-Image~\cite{team2025longcat} & 6B & 0.99 & 0.98 & 0.86 & 0.86 & 0.75 & 0.73 & 0.87 \\
    DeepGen1.0~\cite{wang2026deepgen} & 2B & 0.99 & 0.97 & 0.68 & 0.90 & 0.71 & 0.71 & 0.83 \\
    \midrule
    SANA-1.6B~\cite{xie2024sana} & 1.6B & 0.99 & 0.80 & 0.62 & 0.89 & 0.27 & 0.43 & 0.67 \\
    Hunyuan-DiT~\cite{cao2025hunyuanimage} & 1.5B & 0.97 & 0.77 & 0.71 & 0.88 & 0.13 & 0.30 & 0.63 \\
    MEISSONIC~\cite{bai2024meissonic} & 1B & 0.99 & 0.66 & 0.42 & 0.86 & 0.10 & 0.22 & 0.54 \\
    \midrule
    SnapGen~\cite{hu2024snapgen} & 0.38B & - & - & - & - & - & - & 0.70 \\
    SnapGen++~\cite{hu2026snapgen++} & 0.4B & - & - & - & - & - & - & 0.66 \\
    SANA-0.6B~\cite{xie2024sana} & 0.59B & 0.98 & 0.78 & 0.59 & 0.86 & 0.22 & 0.40 & 0.64 \\
    Nitro-E-GRPO~\cite{shen2025mmdit} & \textbf{0.3B} & \textbf{1.00} & 0.88 & 0.63 & 0.86 & 0.32 & 0.48 & 0.70 \\

    \textbf{Ours} & {0.39B} & {0.98} & \textbf{0.88} & \textbf{0.65} & \textbf{0.90} & \textbf{0.35} & \textbf{0.52} & \textbf{0.72} \\ 
    \bottomrule
    \end{tabular}
}
\end{table*}

\begin{table*}[t]
\centering
\caption{
Evaluation on DPG benchmark. All images are generated at $1024 \times 1024$ resolution. ``Param.'' denotes backbone parameters.
}
\label{tab:dpg_results}

\resizebox{\textwidth}{!}{
    \setlength{\tabcolsep}{12pt} 
    \begin{tabular}{lc|ccccc|c}
    \toprule
    % \multirow{2}{*}{\textbf{Model}} & \textbf{Backbone} & \multicolumn{7}{c}{\textbf{GenEval}} \\ & \textbf{Params} & Single Obj. & Two Obj. & Counting & Colors & Position & Color Attr. & Overall \\ 
    {Model} & {Param.} & Global & Entity & Attribute & Relation & Other & Overall \\
    \midrule
    FLUX.1-Dev~\cite{flux2023} & 12B & 82.1 & 89.5 & 88.7 & 91.1 & 89.4 & 84.0 \\
    BAGEL~\cite{deng2025bagel} & 7B & 88.9 & 90.4 & 91.3 & 90.8 & 88.7 & 85.1 \\
    OmniGen2~\cite{wu2025omnigen2} & 4B & 88.8 & 88.8 & 90.2 & 89.4 & 90.3 & 83.6 \\
    Z-Image~\cite{cai2025z} & 6B & 93.4 & 91.2 & 93.2 & 92.2 & 91.5 & 88.1 \\
    LongCat-Image~\cite{team2025longcat} & 6B & 89.1 & 92.5 & 92.0 & 93.3 & 87.5 & 86.6 \\
    DeepGen1.0~\cite{wang2026deepgen} & 2B & 84.2 & 91.0 & 88.6 & 94.0 & 85.3 & 84.6 \\
    \midrule
    SANA-1.6B~\cite{xie2024sana} & 1.6B & 86.0 & 91.5 & 88.9 & 91.9 & 90.7 & 84.8 \\
    Hunyuan-DiT~\cite{cao2025hunyuanimage} & 1.5B & 84.6 & 80.6 & 88.0 & 74.4 & 86.4 & 78.9 \\
    MEISSONIC~\cite{bai2024meissonic} & 1B & 75.8 & 74.3 & 74.4 & 81.5 & 57.6 & 65.3 \\
    \midrule
    SnapGen~\cite{hu2024snapgen} & 0.38B & - & - & - & - & - & 81.1 \\
    SnapGen++~\cite{hu2026snapgen++} & 0.4B & - & - & - & - & - & 85.2 \\
    SANA-0.6B~\cite{xie2024sana} & 0.59B & 83.0 & 89.5 & 89.3 & 90.1 & \textbf{90.2} & 83.6  \\
    Nitro-E-GRPO~\cite{shen2025mmdit} & \textbf{0.3B} & 82.7 & 87.6 & 86.9 & 92.1 & 74.9 & 81.1 \\

    \textbf{Ours} & {0.39B} & \textbf{84.9} & \textbf{91.6} & \textbf{89.7} & \textbf{94.3} & {80.4} & \textbf{85.8}  \\ 
    \bottomrule
    \end{tabular}
}
\end{table*}

\noindent\textbf{Performance on Image Generation.}
We first assess the generative quality using GenEval~\cite{ghosh2023geneval} and DPG~\cite{hu2024ella} as the primary benchmarks with all images generated at $1024 \times 1024$ resolution for a fair comparison.
The compared baselines are categorized into three groups:
\begin{itemize}
    \item {Unified Models} (\eg, BAGEL, OmniGen2), which typically feature backbone sizes exceeding 2B parameters;
    \item {Lightweight Generative Models ($<2B$)} (\eg, SANA-1.6B, Meissonic), which remain unsuitable for seamless on-device deployment;
    \item {On-device Generative Models ($<1B$)} (\eg, SnapGen series, SANA-0.6B), which are restricted to only generation tasks.
\end{itemize}

Table~\ref{tab:geneval_results} and Table~\ref{tab:dpg_results} details the parameter counts, per-category scores and overall performance on GenEval and DPG benchmark. 
Note that for the SnapGen series, per-category scores are omitted as the models are not open-sourced.
Nevertheless, \ours (0.39B) achieves competitive performance, rivaling models with nearly $10\times$ more parameters.

\begin{table*}[t]
\centering
\caption{
Evaluation on ImgEdit benchmark. ``Param.'' denotes backbone parameters.
}
\label{tab:editing_results}
\small
\resizebox{\textwidth}{!}{
    \begin{tabular}{lc|ccccccccc|c}
    \toprule
    {Model} & {Param.} & Add & Remove & Replace & Adjust & BG & Style & Extract & Action & Hybrid & Overall \\
    \midrule
    % Instruct p2p & - & 2.45 & 1.50 & 2.01 & 1.83 & 1.44 & 3.55 & 1.44 & 1.46 & 1.20 & 1.88 \\
    % MagicBrush & - & 2.84 & 1.58 & 1.97 & 1.58 & 1.75 & 2.38 & 1.51 & 1.22 & 1.62 & 1.83 \\
    % AnyEdit & - & 3.18 & 2.23 & 2.47 & 2.95 & 2.24 & 2.85 & 1.88 & 2.65 & 1.56 & 2.45 \\
    % UltraEdit & - & 3.44 & 1.45 & 2.96 & 2.81 & 2.83 & 3.76 & 2.13 & 2.98 & 1.91 & 2.70 \\
    % ICEdit & - & 3.58 & 2.93 & 3.15 & 3.39 & 3.08 & 3.84 & 1.73 & 3.68 & 2.04 & 3.05 \\
    % Step1X-Edit-v1.1 & - & 3.88 & 2.41 & 3.40 & 3.14 & 3.16 & 4.63 & 1.76 & 2.52 & 2.64 & 3.06 \\
    % UniWorld-V1 & - & 3.82 & 3.24 & 3.47 & 3.64 & 2.99 & 4.21 & 2.27 & 2.74 & 2.96 & 3.26 \\
    Kontext-Dev~\cite{labs2025flux} & 12B & 4.04 & 2.92 & 4.28 & 3.89 & 4.00 & 4.43 & 2.71 & 4.52 & 3.08 & 3.76 \\
    BAGEL~\cite{deng2025bagel} & 7B & 3.81 & 3.16 & 3.85 & 3.59 & 3.39 & 4.51 & 1.58 & 4.25 & 2.67 & 3.42 \\
    OmniGen2~\cite{wu2025omnigen2} & 4B & 3.57 & 3.20 & 3.74 & 3.06 & 3.57 & 4.81 & 1.77 & 4.68 & 2.52 & 3.44 \\
    Z-Image~\cite{cai2025z} & 6B & 4.40 & 4.13 & 4.57 & 4.14 & 4.14 & 4.85 & 4.30 & 4.50 & 3.63 & 4.30 \\
    LongCat-Image~\cite{team2025longcat} & 6B & 4.66 & 4.68 & 4.75 & 4.64 & 4.35 & 4.81 & 3.90 & 4.83 & 3.80 & 4.49 \\
    DeepGen1.0~\cite{wang2026deepgen} & 2B & 4.52 & 4.43 & 4.42 & 4.09 & 4.38 & 4.48 & 1.75 & 4.57 & 3.61 & 4.03 \\
    \midrule
    VIBE~\cite{alekseenko2026vibe} & 1.6B & 3.89 & {4.42} & {4.34} & 4.22 & 4.22 & 4.40 & {2.90} & 2.75 & {3.52} & 3.85  \\
    \midrule
    EditMGT~\cite{chow2025editmgt} & 0.96B & 2.78 & 2.89 & 3.34 & 3.29 & 3.05 & 4.17 & 1.56 & 2.94 & 2.01 & 2.89 \\
    \textbf{Ours} & \textbf{0.39B} & \textbf{4.59} & \textbf{4.27} & \textbf{4.32} & \textbf{4.46} & \textbf{4.28} & \textbf{4.42} & \textbf{2.70} & \textbf{4.73} & \textbf{3.24} & \textbf{4.11} \\
    \bottomrule
    \end{tabular}
}
\end{table*}

\begin{table}[t]
\centering
\caption{
Evaluation on GEdit-EN benchmark. ``Param.'' denotes backbone parameters. $Q_{SC}$, $Q_{PQ}$, and $Q_{O}$ denote Semantic Consistency, Perceptual Quality and Overall score evaluated by Qwen-VL-2.5 respectively.
}
\label{tab:gedit_results}
\small
\resizebox{0.5\linewidth}{!}{
    \begin{tabular}{lc|ccc}
    \toprule
    {Model} & {Param.} & $Q_{SC}$ & $Q_{PQ}$ & $Q_{O}$ \\
    \midrule
    Kontext-Dev~\cite{labs2025flux} & 12B & 6.61 & 7.24 & 6.79 \\
    BAGEL~\cite{deng2025bagel} & 7B & 7.66 & 7.08 & 7.20 \\
    OmniGen2~\cite{wu2025omnigen2} & 4B & 7.11 & 7.22 & 6.79 \\
    % Z-Image~\cite{cai2025z} & 6B & - & - & -  \\
    LongCat-Image~\cite{team2025longcat} & 6B & 7.87 & 7.49 & 7.55 \\
    DeepGen1.0~\cite{wang2026deepgen} & 2B & 7.83 & 7.47 & 7.54 \\
    \midrule
    VIBE~\cite{alekseenko2026vibe} & 1.6B & {7.61} & 7.34 & 7.28 \\
    \midrule
    EditMGT~\cite{chow2025editmgt} & 0.96B & 6.77 & 6.74 & 6.33 \\
    \textbf{Ours} & \textbf{0.39B} & \textbf{7.04} & \textbf{7.54} & \textbf{6.88} \\
    \bottomrule
    \end{tabular}
}
\end{table}

\noindent\textbf{Performance on Image Editing.}
We evaluate editing proficiency using the ImgEdit~\cite{ye2025imgedit} and GEdit~\cite{liu2025step1x-edit} benchmarks.
For ImgEdit, we employ GPT-4o (2024-11-20) as the automated evaluation metric. For GEdit, we focus on English-based editing tasks and utilize Qwen2.5-VL~\cite{bai2025qwen3} as the model-based judge.
Similar to generation task, our method is compared against a broad spectrum of state-of-the-art (SOTA) baselines, categorized into:
\begin{itemize}
    \item {Unified Models} (\eg, FLUX, BAGEL, LongCat-Image);
    \item {Lightweight Editing Models ($<2B$)} (\ie, VIBE);
    \item {On-device Editing Models ($<1B$)} (\ie, EditMGT).
\end{itemize}
Crucially, to the best of our knowledge, \ours is among the first to successfully enable image editing tasks on-device with a model scale significantly under 0.5B parameters.
Table~\ref{tab:editing_results} summarizes the model sizes, per-category scores and overall performance on the ImgEdit benchmark.
\ours achieves SOTA among all lightweight models.

\subsection{Qualitative Results}

To further validate the visual fidelity and instruction-following capability of our \ours, we present a qualitative comparison with both large unified models and specialized lightweight baselines.

\begin{figure*}[!p]
\centering
\includegraphics[width=\textwidth, height=\textheight, keepaspectratio]{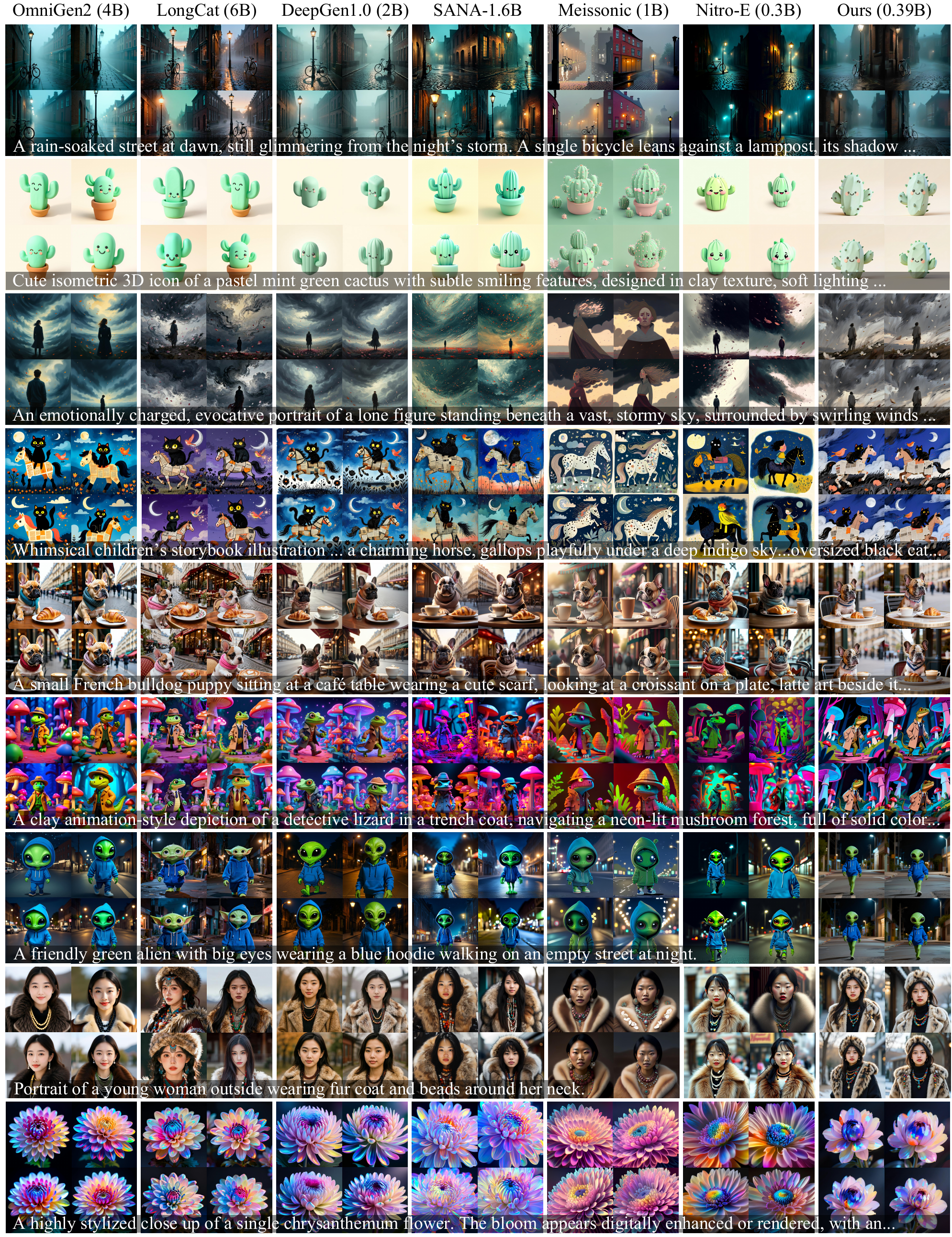}
% \vspace{-20pt}
\caption{Comparison of image generation. All input prompts are listed at the bottom of the images.}
\label{fig:open_gen}
\end{figure*}

\noindent\textbf{Visual Analysis of Image Generation.}
As illustrated in Fig.~\ref{fig:open_gen}, although \ours is significantly smaller than other large-scale models, it maintains high structural integrity and strong semantic alignment across diverse generation scenarios. 
For instance, in complex realistic prompts such as ``A rain-soaked street at dawn...'', \ours accurately captures the atmospheric lighting and produces consistent structural details of the ``bicycle'', avoiding the distortion often seen in smaller counterparts. 
Furthermore, in stylized and imaginative generation tasks, ranging from a ``Cute isometric 3D icon of a pastel mint green cactus in clay texture'' to a ``Whimsical children's storybook illustration'' featuring multiple subjects (a horse and an oversized black cat), our model demonstrates a profound understanding of stylistic nuances and complex spatial compositions.
Compared to other lightweight models like Nitro-E (0.3B) or Meissonic (1B), our results consistently exhibit superior scene composition, better adherence to stylistic constraints (\eg, clay animation-style or stylized digital blooms), and significantly fewer artifacts in high-resolution synthesis. 
While there might be a marginal gap in rendering extremely fine-grained textures in human portraits when compared to larger specialized models like SANA-1.6B, our model provides a more robust and coherent interpretation of environmental lighting, mood, and complex multi-object interactions (\eg, ``A small French bulldog puppy sitting at a café table... looking at a croissant'').

\begin{figure*}[!p]
\centering
\includegraphics[width=\textwidth, height=\textheight, keepaspectratio]{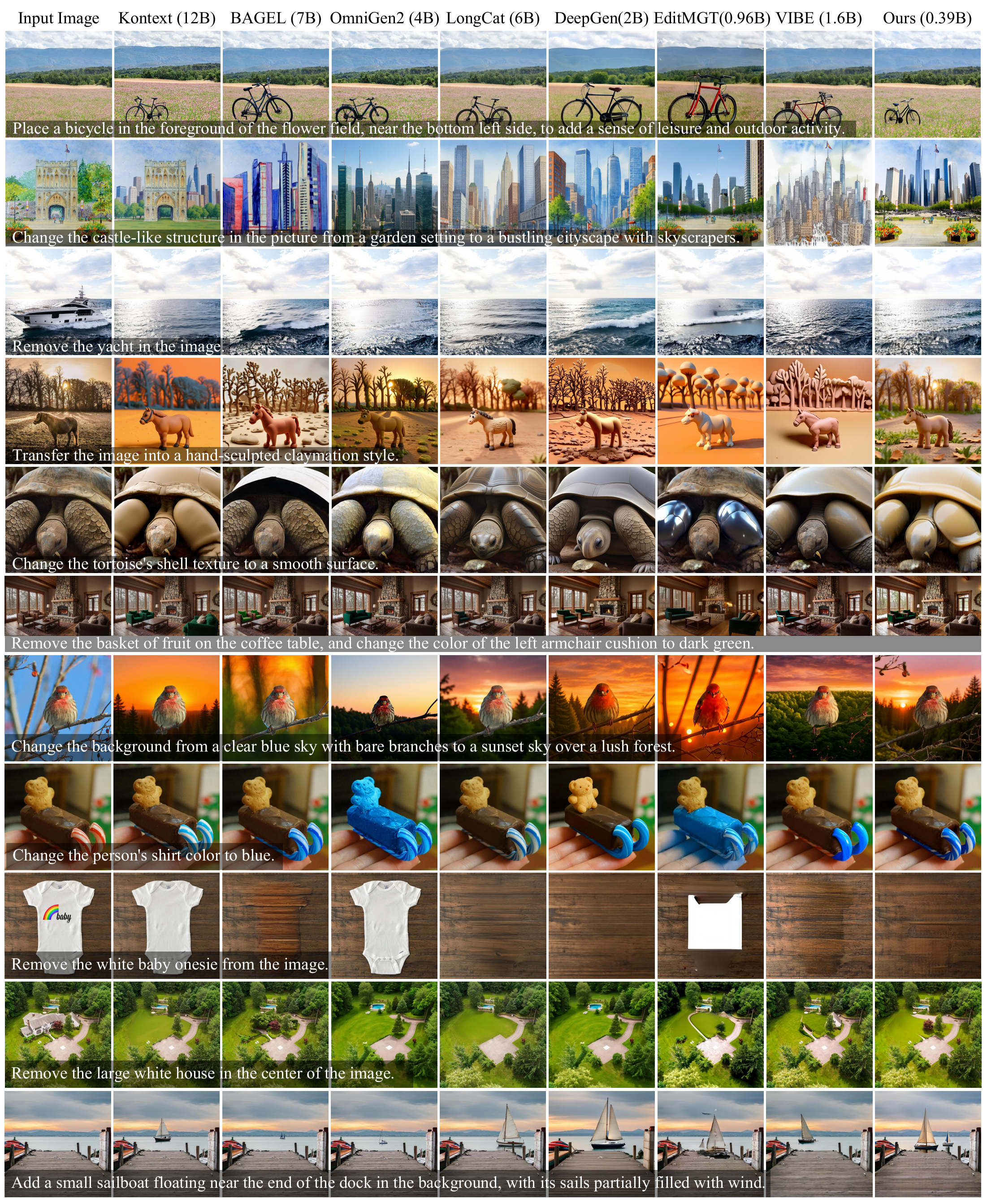}
\caption{Comparison of image editing. All source images and instructions are from ImgEdit Benchmark.}
\label{fig:open_edit}
\end{figure*}

\noindent\textbf{Visual Analysis of Image Editing.}
The image editing comparisons in Fig.~\ref{fig:open_edit} further highlight the performance of \ours. 
For instance, in object addition and removal tasks 
(\eg, ``Place a bicycle in the foreground of the flower field'' or ``Remove the white baby onesie from the image''), 
our model demonstrates a precise understanding of spatial relationships and occlusion comparable to large-scale models (\eg, OmniGen2 and BAGEL). 
Furthermore, when dealing with complex style and background transformations 
(\eg, ``Transfer the image into a hand-sculpted claymation style'' or ``Change the background from a clear blue sky with bare branches to a sunset sky over a lush forest''),
our approach successfully preserves the structural integrity of the main subjects while seamlessly applying the desired modifications.
In addition, for fine-grained localized edits, such as texture alterations and multi-object manipulation
(\eg, ``Change the tortoise's shell texture to a smooth surface'' or ``Remove the basket of fruit on the coffee table, and change the color of the left armchair cushion to dark green''),
our model efficiently executes intricate instructions without unintended disruptions to the surrounding context, proving its high instruction-following capability despite having significantly fewer parameters (0.39B) compared to models like Kontext (12B) or LongCat (6B).

\clearpage

\clearpage

% \subsection{Human Preference Study}

\subsection{Ablation Study}

\begin{table*}[t]
    \small
    \centering
    \caption{Ablation study on GenEval and ImgEdit benchmarks. ``TPJ'' denotes ``Task-progressive Joint''.}
    \label{tab:ablation_paradigm}
    \resizebox{\textwidth}{!}{
        \setlength{\tabcolsep}{8pt} 
        \begin{tabular}{lllcc}
        \toprule
        \textbf{Exp.} & \textbf{Mechanism} & \textbf{Training Stage} & \textbf{GenEval $\uparrow$} & \textbf{ImgEdit $\uparrow$} \\
        \midrule
         \multicolumn{3}{l}{Text-to-image Pretraining} & 0.70  & - \\
        \midrule
        \multirow{2}{100pt}{Condition Mechanism} 
            & Pix2Pix & T2I $\rightarrow$ Edit & 0.56 & 3.67 \\
            & Pix2Pix & T2I $\rightarrow$ Edit $\rightarrow$ Unified & 0.61 & 3.65 \\
        \midrule
        \multirow{4}{100pt}{Training Recipe} 
            & In-context & T2I $\rightarrow$ T2I & 0.65 & - \\
            & In-context & T2I $\rightarrow$ Edit & 0.64 & 3.88 \\
            & In-context & T2I $\rightarrow$ Unified & 0.65  & 3.14 \\
            & In-context & T2I $\rightarrow$ Edit $\rightarrow$ Unified & 0.71 & 3.94 \\
        \midrule
        \multirow{1}{100pt}{Reinforcement Learning}
            & In-context & TPJ Pretrain $\rightarrow$ RLHF & \textbf{0.72} & \textbf{4.11} \\
        \midrule
        \multirow{1}{100pt}{Step Distillation}
            & In-context & TPJ Pretrain $\rightarrow$ RLHF $\rightarrow$ DMD & 0.70 & 3.8  \\
        \bottomrule
        \end{tabular}
    }
\end{table*}

\begin{figure}[t]
    \centering
    \includegraphics[width=\linewidth]{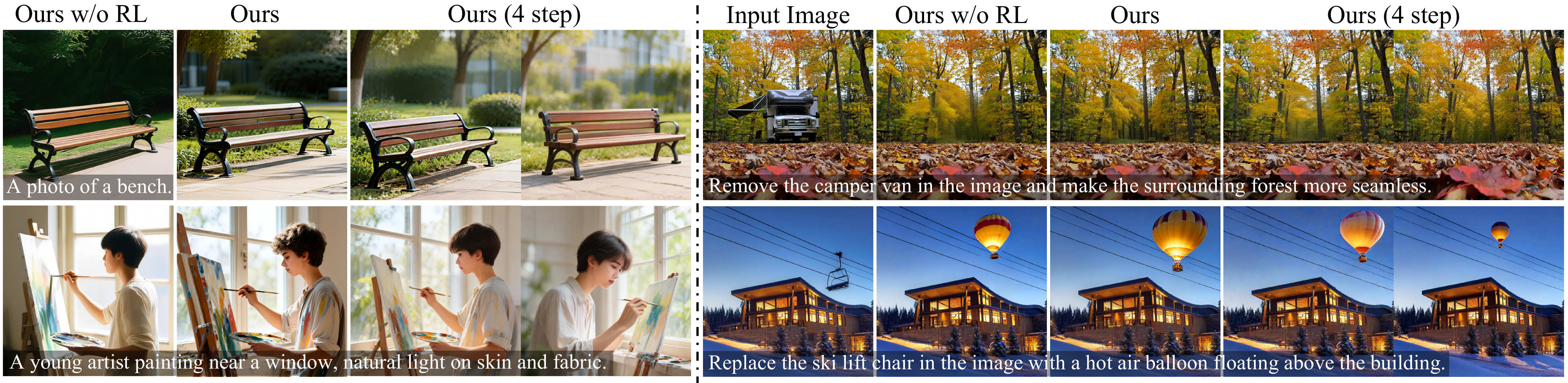}
    \caption{Ablation study on generation (left) and editing (right) tasks. From left to right in each task: results without RL, our full model, and our model after 4-step distillation.}
    \label{fig:ablation}
\end{figure}

To dissect the contribution of our proposed training strategies, we conduct an extensive ablation study on \ours, evaluating generation and editing performance on GenEval and ImgEdit benchmarks.
The quantitative results are summarized in Table~\ref{tab:ablation_paradigm} and the qualitative results are shown in Fig.~\ref{fig:ablation}.

\noindent\textbf{Condition Mechanism.}
We first investigate the effectiveness of the In-context (IC) mechanism compared to the traditional Pix2Pix-style channel concatenation after Text-to-image pretraining.
As shown in Table~\ref{tab:ablation_paradigm}, when maintaining the same editing training stage (\ie, T2I $\rightarrow$ Edit, row 2\&5), the IC mechanism yields a significant improvement from 3.67 to 3.88.
Furthermore, compared to the results at unified training stage (\ie, T2I $\rightarrow$ Edit $\rightarrow$ Unified, row 3\&7), the IC mechanism proves more suitable for joint training, particularly for the generation task (from 0.61 to 0.71).
We attribute this to the fact that IC treats the generation as a special case of in-context referencing (\ie, the condition image is represented by a blank image).
This formulation allows the model to leverage a unified spatial prior across disparate behaviors.

\noindent\textbf{Training Recipe.}
The transition from specialized tasks to unified behavior often encounters a performance bottleneck. Our results indicate that direct joint training (\ie, T2I $\rightarrow$ Unified, row 6) sacrifices both generation and editing precision simultaneously (\eg, GenEval 0.65 and ImgEdit 3.14).
By introducing Task-progressive Joint (TPJ) Pre-training (\ie, T2I $\rightarrow$ Edit $\rightarrow$ Unified, row 7), we observe that the model not only recovers but also surpasses the performance of separate tasks (GenEval 0.71 \emph{vs.} 0.70 and ImgEdit 3.91 \emph{vs.} 3.88).
This demonstrates that TPJ allows the compact backbone to internalize the fundamental logic of IC referencing before tackling unified modeling.
We also conduct in-context T2I training following standard T2I pre-training (row 4), which results in a slight decline in GenEval performance. 
We posit that since the blank image provides no information, the model fails to utilize it for contextual referencing, thereby slowing down convergence and compromising T2I performance.

\noindent\textbf{Reinforcement Learning.}
% Further refinements are explored to enhance both perceptual quality and inference efficiency.
% 
As shown in Table~\ref{tab:ablation_paradigm}, integrating RLHF into the TPJ pipeline yields the highest overall performance, achieving 4.11 on ImgEdit and 0.72 on GenEval.
Beyond these numerical gains, visual evidence in Fig.~\ref{fig:ablation} demonstrates a substantial leap in image aesthetics and high-frequency details.
Specifically, it markedly improves background realism in generation tasks and enhances human identity maintenance during editing.

\noindent\textbf{Step Distillation.}
DMD significantly accelerates the sampling process, albeit with a slight performance penalty in complex semantic alignment tasks.
Fig.~\ref{fig:ablation} presents the visual results of \ours with 4-step inference after Distribution Matching Distillation.
This performance-efficiency trade-off is expected, as the compression of the ODE trajectory inherently constrains the model's capacity to navigate high-dimensional latent manifolds for intricate edits.

\subsection{On-device Deployment}
\label{sec:deployment}
\begin{figure}[h]
    \centering
    \includegraphics[width=\linewidth]{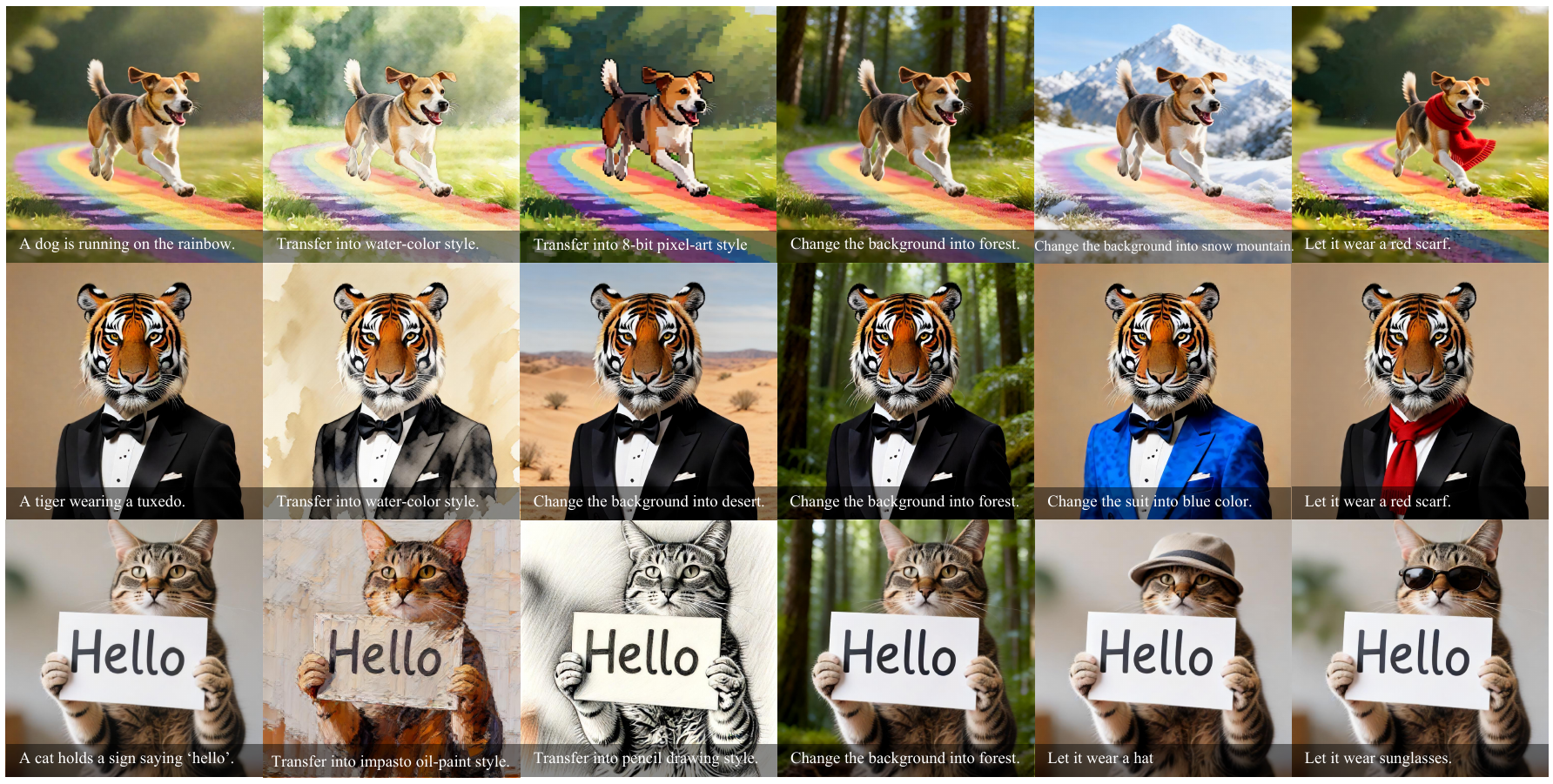}
    \caption{\ours generation and editing examples on mobile device.}
    \label{fig:suppl_mobile}
\end{figure}

To evaluate the practical utility of \ours, we conduct comprehensive experiments on real mobile phones (\ie, Xiaomi 14 and vivo X100), equipped with the Snapdragon 8 Gen3 (Qualcomm NPU) and Dimensity 9300 (MTK APU).
All measurements are performed using a quantized W8A8 (\ie, weights 8-bit, activations 8-bit) U-Net backbone at a $1024 \times 1024$ resolution with a 4-step sampling schedule.

\noindent\textbf{Runtime Optimization.} 
Given that the standard Qwen3-VL text encoder comprises approximately 2B parameters, direct on-device text encoding remains a primary latency bottleneck, 
% with estimated runtimes exceeding 24s per prompt (\ie, 4s for T2I and 24s for I2I). 
% 
We choose to pre-deploy common prompts (\eg, stylized or predefined editing tasks) as pre-computed embeddings for instantaneous interaction on mobile devices.
For future iterations, we are developing a lightweight text encoder (<1B) to enable full-pipeline on-device flexibility without compromising inference speed.

\noindent\textbf{Performance Analysis.} 
As summarized in Table~\ref{tab:on_device_speed}, the U-Net inference per step is merely 103.84ms on the Snapdragon 8 Gen3, leading to a total generation or editing time of approximately {0.42s} (excluding VAE). 
Including VAE decoding ($\sim$22ms) and system overhead, the end-to-end user experience remains near the 1s threshold. 
% 
% These results validate \ours as a highly viable solution for real-time generative applications on modern smartphones. 
Fig.~\ref{fig:suppl_mobile} shows our generation (left) and editing (right) results. The UNet is quantized as W8A8 with a 4-step inference. 

\begin{table}[t]
\centering
\caption{On-device inference latency (ms) for \ours components at $1024 \times 1024$ resolution. Measurements are conducted on W8A8 quantized models.}
\label{tab:on_device_speed}
\resizebox{0.8\columnwidth}{!}{
    \setlength{\tabcolsep}{5pt}
    \begin{tabular}{l|cc|c}
    \toprule
    \textbf{Component} & \textbf{Snapdragon 8 Gen3} & \textbf{Dimensity 9300} & \textbf{Model Size} \\
    \midrule
    VAE Encoder (fp16) & 22.17 & 26.16 & 2.45 MB \\
    U-Net (w8a8, per step) & 103.84 & 122.53 & 389.00 MB \\
    VAE Decoder (fp16) & 22.17 & 26.16 & 2.45 MB \\
    \midrule
    {Total (4 steps, U-Net)} & {415.36} & {490.12} & {393.90 MB} \\
    \bottomrule
    \end{tabular}
}
\end{table}

\section{Limitations}
\label{sec:limitations}

Despite the efficiency and versatility of \ours, several limitations remain that define our future research directions. 

\noindent\textbf{Text Encoder Scale.} Although our U-Net backbone is highly compact (0.39B), the current pipeline still relies on a standard 2B-parameter text encoder. During on-device deployment, this component introduces non-negligible latency. 
To achieve a fully optimized on-device pipeline, we are actively developing a lightweight text encoder (<1B) to ensure seamless end-to-end inference.

\noindent\textbf{Reconstruction Fidelity and Specialized Tasks.} 
While \ours performs competitively on major benchmarks, we observe relatively lower scores on GEdit and certain qualitative artifacts in challenging scenarios (\ie, text generation, text editing and identity preservation in portrait editing). 
We attribute these bottlenecks to our extremely compact VAE (1.2M), which may inevitably suffer from information loss or reconstruction blurriness when handling complex structural details. 
To mitigate this, we plan to train a slightly larger, high-fidelity VAE. 
Furthermore, we are going to implement specialized fine-tuning for text and facial generation or editing to enhance perceptual quality and instruction following.

\noindent\textbf{Multi-modal Alignment.} 
Future work will also explore a more advanced step distillation scheme combined with reward model during the post training stage to better align our compact model with human aesthetic preferences and complex spatial instructions.

\section{Conclusion}

We presented \ours, a compact unified on-device diffusion model that supports both text-to-image generation and text-guided image editing within a single network.
%
% \ours unifies conditioning through in-context spatial concatenation in latent space, together with explicit task tokens for disambiguation. 
By unifying conditioning through in-context spatial concatenation and explicit task tokens, \ours achieves seamless multi-tasking without parameter overhead.
To stably train this small-capacity unified model, we introduced a task-progressive curriculum (T2I → Edit → Joint) followed by high-quality SFT and reinforcement learning. 
Finally, step distillation compresses sampling to 4 denoising steps for efficient mobile inference. 
%
% Experiments show that \ours achieves strong performance on standard benchmarks (\eg, GenEval 0.72, ImgEdit 4.11), outperforming prior on-device baselines while remaining competitive with much larger server-side models.
Experiments show that \ours (GenEval 0.72, ImgEdit 4.11) significantly outperforms prior on-device baselines and reamins competitive with much larger server-side model.
On representative smartphone (\ie, Xiaomi 14), \ours could generate or edit a 1024 $\times$ 1024 image in less than 1s.

\clearpage

\bibliographystyle{plainnat}
\bibliography{main}

@article{hu2024snapgen,
  title={Snapgen: Taming high-resolution text-to-image models for mobile devices with efficient architectures and training},
  author={Hu, Dongting and Chen, Jierun and Huang, Xijie and Coskun, Huseyin and Sahni, Arpit and Gupta, Aarush and Goyal, Anujraaj and Lahiri, Dishani and Singh, Rajesh and Idelbayev, Yerlan and others},
  journal={arXiv preprint arXiv:2412.09619},
  year={2024}
}

@article{hu2026snapgen++,
  title={SnapGen++: Unleashing Diffusion Transformers for Efficient High-Fidelity Image Generation on Edge Devices},
  author={Hu, Dongting and Gupta, Aarush and Gabidolla, Magzhan and Sahni, Arpit and Coskun, Huseyin and Li, Yanyu and Idelbayev, Yerlan and Mahmood, Ahsan and Lebedev, Aleksei and Lahiri, Dishani and others},
  journal={arXiv preprint arXiv:2601.08303},
  year={2026}
}

@article{cao2025hunyuanimage,
  title={Hunyuanimage 3.0 technical report},
  author={Cao, Siyu and Chen, Hangting and Chen, Peng and Cheng, Yiji and Cui, Yutao and Deng, Xinchi and Dong, Ying and Gong, Kipper and Gu, Tianpeng and Gu, Xiusen and others},
  journal={arXiv preprint arXiv:2509.23951},
  year={2025}
}

@article{cai2025z,
  title={Z-image: An efficient image generation foundation model with single-stream diffusion transformer},
  author={Cai, Huanqia and Cao, Sihan and Du, Ruoyi and Gao, Peng and Hoi, Steven and Hou, Zhaohui and Huang, Shijie and Jiang, Dengyang and Jin, Xin and Li, Liangchen and others},
  journal={arXiv preprint arXiv:2511.22699},
  year={2025}
}

@article{seedream2025seedream,
  title={Seedream 4.0: Toward next-generation multimodal image generation},
  author={Seedream, Team and Chen, Yunpeng and Gao, Yu and Gong, Lixue and Guo, Meng and Guo, Qiushan and Guo, Zhiyao and Hou, Xiaoxia and Huang, Weilin and Huang, Yixuan and others},
  journal={arXiv preprint arXiv:2509.20427},
  year={2025}
}

@article{gao2025seedream,
  title={Seedream 3.0 technical report},
  author={Gao, Yu and Gong, Lixue and Guo, Qiushan and Hou, Xiaoxia and Lai, Zhichao and Li, Fanshi and Li, Liang and Lian, Xiaochen and Liao, Chao and Liu, Liyang and others},
  journal={arXiv preprint arXiv:2504.11346},
  year={2025}
}

@article{gong2025seedream,
  title={Seedream 2.0: A native chinese-english bilingual image generation foundation model},
  author={Gong, Lixue and Hou, Xiaoxia and Li, Fanshi and Li, Liang and Lian, Xiaochen and Liu, Fei and Liu, Liyang and Liu, Wei and Lu, Wei and Shi, Yichun and others},
  journal={arXiv preprint arXiv:2503.07703},
  year={2025}
}

@misc{flux2023,
  author={Black Forest Labs},
  title={FLUX},
  year={2024},
  howpublished={\url{https://github.com/black-forest-labs/flux}},
}

@misc{flux2025,
  author={Black Forest Labs},
  title={FLUX2},
  year={2025},
  howpublished={\url{https://github.com/black-forest-labs/flux2}},
}

@article{labs2025flux,
  title={FLUX. 1 Kontext: Flow Matching for In-Context Image Generation and Editing in Latent Space},
  author={Labs, Black Forest and Batifol, Stephen and Blattmann, Andreas and Boesel, Frederic and Consul, Saksham and Diagne, Cyril and Dockhorn, Tim and English, Jack and English, Zion and Esser, Patrick and others},
  journal={arXiv preprint arXiv:2506.15742},
  year={2025}
}

@misc{taesd2024,
  author={Madebyollin},
  title={taesd},
  year={2024},
  howpublished={\url{https://github.com/madebyollin/taesd}},
}

@article{wu2025qwen,
  title={Qwen-image technical report},
  author={Wu, Chenfei and Li, Jiahao and Zhou, Jingren and Lin, Junyang and Gao, Kaiyuan and Yan, Kun and Yin, Sheng-ming and Bai, Shuai and Xu, Xiao and Chen, Yilei and others},
  journal={arXiv preprint arXiv:2508.02324},
  year={2025}
}

@article{xie2024sana,
  title={Sana: Efficient high-resolution image synthesis with linear diffusion transformers},
  author={Xie, Enze and Chen, Junsong and Chen, Junyu and Cai, Han and Tang, Haotian and Lin, Yujun and Zhang, Zhekai and Li, Muyang and Zhu, Ligeng and Lu, Yao and others},
  journal={arXiv preprint arXiv:2410.10629},
  year={2024}
}

@article{wang2026deepgen,
  title={DeepGen 1.0: A Lightweight Unified Multimodal Model for Advancing Image Generation and Editing},
  author={Wang, Dianyi and Li, Ruihang and Han, Feng and Ma, Chaofan and Song, Wei and Wang, Siyuan and Wang, Yibin and Xin, Yi and Liu, Hongjian and Zhang, Zhixiong and others},
  journal={arXiv preprint arXiv:2602.12205},
  year={2026}
}

@article{alekseenko2026vibe,
  title={VIBE: Visual Instruction Based Editor},
  author={Alekseenko, Grigorii and Gordeev, Aleksandr and Tolstykh, Irina and Suleimanov, Bulat and Dokholyan, Vladimir and Fedorov, Georgii and Yakubson, Sergey and Tsybina, Aleksandra and Chernyshov, Mikhail and Kuprashevich, Maksim},
  journal={arXiv preprint arXiv:2601.02242},
  year={2026}
}

@article{li2023snapfusion,
  title={Snapfusion: Text-to-image diffusion model on mobile devices within two seconds},
  author={Li, Yanyu and Wang, Huan and Jin, Qing and Hu, Ju and Chemerys, Pavlo and Fu, Yun and Wang, Yanzhi and Tulyakov, Sergey and Ren, Jian},
  journal={Advances in Neural Information Processing Systems},
  volume={36},
  pages={20662--20678},
  year={2023}
}

@inproceedings{zhao2024mobilediffusion,
  title={Mobilediffusion: Instant text-to-image generation on mobile devices},
  author={Zhao, Yang and Xu, Yanwu and Xiao, Zhisheng and Jia, Haolin and Hou, Tingbo},
  booktitle={European Conference on Computer Vision},
  pages={225--242},
  year={2024},
  organization={Springer}
}

@inproceedings{brooks2023instructpix2pix,
  title={Instructpix2pix: Learning to follow image editing instructions},
  author={Brooks, Tim and Holynski, Aleksander and Efros, Alexei A},
  booktitle={Proceedings of the IEEE/CVF conference on computer vision and pattern recognition},
  pages={18392--18402},
  year={2023}
}

@inproceedings{ma2025hpsv3,
  title={Hpsv3: Towards wide-spectrum human preference score},
  author={Ma, Yuhang and Wu, Xiaoshi and Sun, Keqiang and Li, Hongsheng},
  booktitle={Proceedings of the IEEE/CVF International Conference on Computer Vision},
  pages={15086--15095},
  year={2025}
}

@article{wu2025editreward,
  title={Editreward: A human-aligned reward model for instruction-guided image editing},
  author={Wu, Keming and Jiang, Sicong and Ku, Max and Nie, Ping and Liu, Minghao and Chen, Wenhu},
  journal={arXiv preprint arXiv:2509.26346},
  year={2025}
}

@article{xu2023imagereward,
  title={Imagereward: Learning and evaluating human preferences for text-to-image generation},
  author={Xu, Jiazheng and Liu, Xiao and Wu, Yuchen and Tong, Yuxuan and Li, Qinkai and Ding, Ming and Tang, Jie and Dong, Yuxiao},
  journal={Advances in Neural Information Processing Systems},
  volume={36},
  pages={15903--15935},
  year={2023}
}

@article{yin2024improved,
  title={Improved distribution matching distillation for fast image synthesis},
  author={Yin, Tianwei and Gharbi, Micha{\"e}l and Park, Taesung and Zhang, Richard and Shechtman, Eli and Durand, Fredo and Freeman, Bill},
  journal={Advances in neural information processing systems},
  volume={37},
  pages={47455--47487},
  year={2024}
}

@inproceedings{yin2024one,
  title={One-step diffusion with distribution matching distillation},
  author={Yin, Tianwei and Gharbi, Micha{\"e}l and Zhang, Richard and Shechtman, Eli and Durand, Fredo and Freeman, William T and Park, Taesung},
  booktitle={Proceedings of the IEEE/CVF conference on computer vision and pattern recognition},
  pages={6613--6623},
  year={2024}
}

@inproceedings{chen2024pixart,
  title={Pixart-$\sigma$: Weak-to-strong training of diffusion transformer for 4k text-to-image generation},
  author={Chen, Junsong and Ge, Chongjian and Xie, Enze and Wu, Yue and Yao, Lewei and Ren, Xiaozhe and Wang, Zhongdao and Luo, Ping and Lu, Huchuan and Li, Zhenguo},
  booktitle={European Conference on Computer Vision},
  pages={74--91},
  year={2024},
  organization={Springer}
}

@article{luo2023latent,
  title={Latent consistency models: Synthesizing high-resolution images with few-step inference},
  author={Luo, Simian and Tan, Yiqin and Huang, Longbo and Li, Jian and Zhao, Hang},
  journal={arXiv preprint arXiv:2310.04378},
  year={2023}
}

@article{podell2023sdxl,
  title={Sdxl: Improving latent diffusion models for high-resolution image synthesis},
  author={Podell, Dustin and English, Zion and Lacey, Kyle and Blattmann, Andreas and Dockhorn, Tim and M{\"u}ller, Jonas and Penna, Joe and Rombach, Robin},
  journal={arXiv preprint arXiv:2307.01952},
  year={2023}
}

@inproceedings{rombach2022high,
  title={High-resolution image synthesis with latent diffusion models},
  author={Rombach, Robin and Blattmann, Andreas and Lorenz, Dominik and Esser, Patrick and Ommer, Bj{\"o}rn},
  booktitle={Proceedings of the IEEE/CVF conference on computer vision and pattern recognition},
  pages={10684--10695},
  year={2022}
}

@article{bai2025qwen3,
  title={Qwen3-vl technical report},
  author={Bai, Shuai and Cai, Yuxuan and Chen, Ruizhe and Chen, Keqin and Chen, Xionghui and Cheng, Zesen and Deng, Lianghao and Ding, Wei and Gao, Chang and Ge, Chunjiang and others},
  journal={arXiv preprint arXiv:2511.21631},
  year={2025}
}

@article{lin2025uniworld,
  title={Uniworld-v1: High-resolution semantic encoders for unified visual understanding and generation},
  author={Lin, Bin and Li, Zongjian and Cheng, Xinhua and Niu, Yuwei and Ye, Yang and He, Xianyi and Yuan, Shenghai and Yu, Wangbo and Wang, Shaodong and Ge, Yunyang and others},
  journal={arXiv preprint arXiv:2506.03147},
  year={2025}
}

@article{liu2025flow,
  title={Flow-grpo: Training flow matching models via online rl},
  author={Liu, Jie and Liu, Gongye and Liang, Jiajun and Li, Yangguang and Liu, Jiaheng and Wang, Xintao and Wan, Pengfei and Zhang, Di and Ouyang, Wanli},
  journal={arXiv preprint arXiv:2505.05470},
  year={2025}
}

@article{xue2025dancegrpo,
  title={Dancegrpo: Unleashing grpo on visual generation},
  author={Xue, Zeyue and Wu, Jie and Gao, Yu and Kong, Fangyuan and Zhu, Lingting and Chen, Mengzhao and Liu, Zhiheng and Liu, Wei and Guo, Qiushan and Huang, Weilin and others},
  journal={arXiv preprint arXiv:2505.07818},
  year={2025}
}

@inproceedings{wallace2024diffusion,
  title={Diffusion model alignment using direct preference optimization},
  author={Wallace, Bram and Dang, Meihua and Rafailov, Rafael and Zhou, Linqi and Lou, Aaron and Purushwalkam, Senthil and Ermon, Stefano and Xiong, Caiming and Joty, Shafiq and Naik, Nikhil},
  booktitle={Proceedings of the IEEE/CVF Conference on Computer Vision and Pattern Recognition},
  pages={8228--8238},
  year={2024}
}

@inproceedings{esser2024scaling,
  title={Scaling rectified flow transformers for high-resolution image synthesis},
  author={Esser, Patrick and Kulal, Sumith and Blattmann, Andreas and Entezari, Rahim and M{\"u}ller, Jonas and Saini, Harry and Levi, Yam and Lorenz, Dominik and Sauer, Axel and Boesel, Frederic and others},
  booktitle={Forty-first international conference on machine learning},
  year={2024}
}

@misc{GPT-Image,
  author={OpenAI},
  title={GPT-Image},
  year={2025},
  howpublished={\url{https://openai.com/index/gpt-4o-system-card/}},
}

@misc{Gemini,
  author={Google},
  title={Gemini 2.5 Flash},
  year={2025},
  howpublished={\url{https://aistudio.google.com/models/gemini-2-5-flash-image}},
}

@article{lipman2022flow,
  title={Flow matching for generative modeling},
  author={Lipman, Yaron and Chen, Ricky TQ and Ben-Hamu, Heli and Nickel, Maximilian and Le, Matt},
  journal={arXiv preprint arXiv:2210.02747},
  year={2022}
}

@article{liu2022flow,
  title={Flow straight and fast: Learning to generate and transfer data with rectified flow},
  author={Liu, Xingchao and Gong, Chengyue and Liu, Qiang},
  journal={arXiv preprint arXiv:2209.03003},
  year={2022}
}

@article{wu2025omnigen2,
  title={Omnigen2: Exploration to advanced multimodal generation},
  author={Wu, Chenyuan and Zheng, Pengfei and Yan, Ruiran and Xiao, Shitao and Luo, Xin and Wang, Yueze and Li, Wanli and Jiang, Xiyan and Liu, Yexin and Zhou, Junjie and others},
  journal={arXiv preprint arXiv:2506.18871},
  year={2025}
}

@article{team2025longcat,
  title={Longcat-image technical report},
  author={Team, Meituan LongCat and Ma, Hanghang and Tan, Haoxian and Huang, Jiale and Wu, Junqiang and He, Jun-Yan and Gao, Lishuai and Xiao, Songlin and Wei, Xiaoming and Ma, Xiaoqi and others},
  journal={arXiv preprint arXiv:2512.07584},
  year={2025}
}

@article{chow2025editmgt,
  title={EditMGT: Unleashing Potentials of Masked Generative Transformers in Image Editing},
  author={Chow, Wei and Li, Linfeng and Kong, Lingdong and Li, Zefeng and Xu, Qi and Song, Hang and Ye, Tian and Wang, Xian and Bai, Jinbin and Xu, Shilin and others},
  journal={arXiv preprint arXiv:2512.11715},
  year={2025}
}

@article{deng2025bagel,
  title   = {Emerging Properties in Unified Multimodal Pretraining},
  author  = {Deng, Chaorui and Zhu, Deyao and Li, Kunchang and Gou, Chenhui and Li, Feng and Wang, Zeyu and Zhong, Shu and Yu, Weihao and Nie, Xiaonan and Song, Ziang and Shi, Guang and Fan, Haoqi},
  journal = {arXiv preprint arXiv:2505.14683},
  year    = {2025}
}

@inproceedings{bai2024meissonic,
  title={Meissonic: Revitalizing masked generative transformers for efficient high-resolution text-to-image synthesis},
  author={Bai, Jinbin and Ye, Tian and Chow, Wei and Song, Enxin and Chen, Qing-Guo and Li, Xiangtai and Dong, Zhen and Zhu, Lei and Yan, Shuicheng},
  booktitle={The Thirteenth International Conference on Learning Representations},
  year={2024}
}

@article{shen2025mmdit,
  title={E-MMDiT: Revisiting Multimodal Diffusion Transformer Design for Fast Image Synthesis under Limited Resources},
  author={Shen, Tong and Yu, Jingai and Zhou, Dong and Li, Dong and Barsoum, Emad},
  journal={arXiv preprint arXiv:2510.27135},
  year={2025}
}

@article{ye2025imgedit,
  title={Imgedit: A unified image editing dataset and benchmark},
  author={Ye, Yang and He, Xianyi and Li, Zongjian and Lin, Bin and Yuan, Shenghai and Yan, Zhiyuan and Hou, Bohan and Yuan, Li},
  journal={arXiv preprint arXiv:2505.20275},
  year={2025}
}

@article{liu2025step1x-edit,
  title={Step1X-Edit: A Practical Framework for General Image Editing}, 
  author={Shiyu Liu and Yucheng Han and Peng Xing and Fukun Yin and Rui Wang and Wei Cheng and Jiaqi Liao and Yingming Wang and Honghao Fu and Chunrui Han and Guopeng Li and Yuang Peng and Quan Sun and Jingwei Wu and Yan Cai and Zheng Ge and Ranchen Ming and Lei Xia and Xianfang Zeng and Yibo Zhu and Binxing Jiao and Xiangyu Zhang and Gang Yu and Daxin Jiang},
  journal={arXiv preprint arXiv:2504.17761},
  year={2025}
}

@article{ghosh2023geneval,
  title={Geneval: An object-focused framework for evaluating text-to-image alignment},
  author={Ghosh, Dhruba and Hajishirzi, Hannaneh and Schmidt, Ludwig},
  journal={Advances in Neural Information Processing Systems},
  volume={36},
  pages={52132--52152},
  year={2023}
}

@article{hu2024ella,
  title={Ella: Equip diffusion models with llm for enhanced semantic alignment},
  author={Hu, Xiwei and Wang, Rui and Fang, Yixiao and Fu, Bin and Cheng, Pei and Yu, Gang},
  journal={arXiv preprint arXiv:2403.05135},
  year={2024}
}

@inproceedings{sauer2024adversarial,
  title={Adversarial diffusion distillation},
  author={Sauer, Axel and Lorenz, Dominik and Blattmann, Andreas and Rombach, Robin},
  booktitle={European Conference on Computer Vision},
  pages={87--103},
  year={2024},
  organization={Springer}
}

@inproceedings{geng2024instructdiffusion,
  title={Instructdiffusion: A generalist modeling interface for vision tasks},
  author={Geng, Zigang and Yang, Binxin and Hang, Tiankai and Li, Chen and Gu, Shuyang and Zhang, Ting and Bao, Jianmin and Zhang, Zheng and Li, Houqiang and Hu, Han and others},
  booktitle={Proceedings of the IEEE/CVF Conference on computer vision and pattern recognition},
  pages={12709--12720},
  year={2024}
}

@inproceedings{huang2024smartedit,
  title={Smartedit: Exploring complex instruction-based image editing with multimodal large language models},
  author={Huang, Yuzhou and Xie, Liangbin and Wang, Xintao and Yuan, Ziyang and Cun, Xiaodong and Ge, Yixiao and Zhou, Jiantao and Dong, Chao and Huang, Rui and Zhang, Ruimao and others},
  booktitle={Proceedings of the IEEE/CVF Conference on Computer Vision and Pattern Recognition},
  pages={8362--8371},
  year={2024}
}

@article{wu2023human,
  title={Human preference score v2: A solid benchmark for evaluating human preferences of text-to-image synthesis},
  author={Wu, Xiaoshi and Hao, Yiming and Sun, Keqiang and Chen, Yixiong and Zhu, Feng and Zhao, Rui and Li, Hongsheng},
  journal={arXiv preprint arXiv:2306.09341},
  year={2023}
}

@article{kirstain2023pick,
  title={Pick-a-pic: An open dataset of user preferences for text-to-image generation},
  author={Kirstain, Yuval and Polyak, Adam and Singer, Uriel and Matiana, Shahbuland and Penna, Joe and Levy, Omer},
  journal={Advances in neural information processing systems},
  volume={36},
  pages={36652--36663},
  year={2023}
}

@article{prabhudesai2023aligning,
  title={Aligning text-to-image diffusion models with reward backpropagation},
  author={Prabhudesai, Mihir and Goyal, Anirudh and Pathak, Deepak and Fragkiadaki, Katerina},
  year={2023}
}

@misc{kakaobrain2022coyo,
  title         = {COYO-700M: Image-Text Pair Dataset},
  author        = {Minwoo Byeon and Beomhee Park and Haecheon Kim and Sungjun Lee and Woonhyuk Baek and Saehoon Kim},
  year          = {2022},
  howpublished  = {\url{https://github.com/kakaobrain/coyo-dataset}},
}

@article{schuhmann2022laion,
  title={Laion-5b: An open large-scale dataset for training next generation image-text models},
  author={Schuhmann, Christoph and Beaumont, Romain and Vencu, Richard and Gordon, Cade and Wightman, Ross and Cherti, Mehdi and Coombes, Theo and Katta, Aarush and Mullis, Clayton and Wortsman, Mitchell and others},
  journal={Advances in neural information processing systems},
  volume={35},
  pages={25278--25294},
  year={2022}
}

@article{sun2023journeydb,
  title={Journeydb: A benchmark for generative image understanding},
  author={Sun, Keqiang and Pan, Junting and Ge, Yuying and Li, Hao and Duan, Haodong and Wu, Xiaoshi and Zhang, Renrui and Zhou, Aojun and Qin, Zipeng and Wang, Yi and others},
  journal={Advances in neural information processing systems},
  volume={36},
  pages={49659--49678},
  year={2023}
}

@article{yue2026know,
  title={Know Your Step: Faster and Better Alignment for Flow Matching Models via Step-aware Advantages},
  author={Yue, Zhixiong and Ni, Zixuan and Ye, Feiyang and Zhang, Jinshan and Shen, Sheng and Mi, Zhenpeng},
  journal={arXiv preprint arXiv:2602.01591},
  year={2026}
}

@inproceedings{jia2025reward,
  title={Reward fine-tuning two-step diffusion models via learning differentiable latent-space surrogate reward},
  author={Jia, Zhiwei and Nan, Yuesong and Zhao, Huixi and Liu, Gengdai},
  booktitle={Proceedings of the Computer Vision and Pattern Recognition Conference},
  pages={12912--12922},
  year={2025}
}

@article{li2024reward,
  title={Reward guided latent consistency distillation},
  author={Li, Jiachen and Feng, Weixi and Chen, Wenhu and Wang, William Yang},
  journal={arXiv preprint arXiv:2403.11027},
  year={2024}
}

@article{luo2024diff,
  title={Diff-instruct++: Training one-step text-to-image generator model to align with human preferences},
  author={Luo, Weijian},
  journal={arXiv preprint arXiv:2410.18881},
  year={2024}
}

@article{luo2025reward,
  title={Reward-instruct: A reward-centric approach to fast photo-realistic image generation},
  author={Luo, Yihong and Hu, Tianyang and Luo, Weijian and Kawaguchi, Kenji and Tang, Jing},
  journal={arXiv preprint arXiv:2503.13070},
  year={2025}
}

@article{shaker2026mobile,
  title={Mobile-O: Unified Multimodal Understanding and Generation on Mobile Device},
  author={Shaker, Abdelrahman and Heakl, Ahmed and Muhammad, Jaseel and Thawkar, Ritesh and Thawakar, Omkar and Li, Senmao and Cholakkal, Hisham and Reid, Ian and Xing, Eric P and Khan, Salman and others},
  journal={arXiv preprint arXiv:2602.20161},
  year={2026}
}

\end{document}